%% file: main.tex
\documentclass{article}

\usepackage[preprint]{neurips_2026}

\usepackage[utf8]{inputenc}
\usepackage[T1]{fontenc}
\usepackage{hyperref}
\usepackage{url}
\usepackage{booktabs}
\usepackage{amsmath}
\usepackage{amsfonts}
\usepackage{nicefrac}
\usepackage{microtype}
\usepackage[table]{xcolor}
\usepackage{graphicx}
\usepackage[accsupp]{axessibility}
\usepackage{multirow}
\usepackage{wrapfig}
\usepackage{caption}
\usepackage{xspace}
\usepackage{bibunits}
\usepackage{subcaption}

\newcommand{\framework}{\texttt{REUNION}\xspace}

\title{Contrastive Hypergraph Source-free Domain Adaptive Object Detection in Adverse Weather Conditions}
\date{}

\author{%
  Jicheng Yuan\textsuperscript{*\ensuremath{\dagger},1}
  \And
  Tung Kieu\textsuperscript{\ensuremath{\dagger},2}
  \And
  Manh Nguyen-Duc\textsuperscript{1}
  \AND
  Manfred Hauswirth\textsuperscript{1,3}
  \And
  Danh Le-Phuoc\textsuperscript{1,3}
  \AND
  \normalfont\small
  \textsuperscript{1}Open Distributed Systems, Technical University of Berlin \\
  \textsuperscript{2}Department of Computer Science, Aalborg University, Denmark \\
  \textsuperscript{3}Fraunhofer Institute for Open Communication Systems, Berlin, Germany \\[2pt]
  \texttt{tungkvt@cs.aau.dk} \\
  \texttt{\{jicheng.yuan,duc.manh.nguyen,} \\
  \texttt{manfred.hauswirth,danh.lephuoc\}@tu-berlin.de}
}

\begin{document}

\maketitle
\begingroup
  \renewcommand{\thefootnote}{\fnsymbol{footnote}}
  \footnotetext[1]{Corresponding author.}
  \footnotetext[2]{Equal contribution.}
\endgroup

\input{00abstract}
\input{01intro}

\input{02related}
\input{03metho}
\input{04exper}
\input{05conclu}

\section{Acknowledgements}
This work is supported by the Deutsche Forschungsgemeinschaft, German Research Foundation under grant number 453130567 (COSMO), by the Horizon Europe Research and Innovation Actions under grant number 101092908 (SmartEdge), by the Chips Joint Undertaking (JU), European Union (EU) HORIZON-JU-IA, under grant agreement No. 101140087 (SMARTY), by the Federal Ministry for Education and Research, Germany under grant number BIFOLD25B.
\vspace{-3mm}

\bibliographystyle{plain}
\bibliography{main}


\newpage
\setcounter{figure}{0}
\renewcommand{\thefigure}{A.\arabic{figure}}
\setcounter{table}{0}
\renewcommand{\thetable}{A.\arabic{table}}
\setcounter{page}{1}
\setcounter{figure}{0}
\setcounter{table}{0}
\renewcommand{\thepage}{S\arabic{page}}
\renewcommand{\thefigure}{A\arabic{figure}}
\renewcommand{\thetable}{A\arabic{table}}
\begin{bibunit}[rusnat]
\input{07appendix}

\putbib[appendix]
\end{bibunit}
\newpage
\setcounter{figure}{0}
\renewcommand{\thefigure}{A.\arabic{figure}}
\setcounter{table}{0}
\renewcommand{\thetable}{A.\arabic{table}}
\setcounter{page}{1}
\setcounter{figure}{0}
\setcounter{table}{0}
\renewcommand{\thepage}{C\arabic{page}}
\renewcommand{\thefigure}{A\arabic{figure}}
\renewcommand{\thetable}{A\arabic{table}}

\end{document}

%% file: 00abstract.tex
\begin{abstract}
Adverse weather conditions such as fog, rain, snow, and low illumination introduce structured visibility degradations that severely impair object detectors trained on clear-weather data. When source data are inaccessible due to privacy or transmission constraints, source-free domain adaptation (SFDA) becomes necessary. However, existing SFDA methods based on pairwise contrastive learning struggle to capture high-order semantics and often suffer from agreement collapse under severe corruption. 
We propose \framework, a hypergraph-guided SFDA framework for object detection in adverse weather conditions. \framework constructs a contrastive hypergraph over target-domain object proposals, encoding high-order relations through intra-image context, weather-aware grouping, prototype-based semantic anchors, and uncertainty-aware connections between reliable and low-confidence instances. To effectively exploit these structured relations, we introduce a group-wise Hyper-InfoNCE objective that optimizes representations at the hyperedge level, enabling semantic information to propagate from confident proposals to corrupted or low-contrast instances. Experiments with diverse benchmarks demonstrate that \framework effectively mitigates domain shifts and achieves state-of-the-art performance in SFDA settings. The code is available at  \href{https://anonymous.4open.science/r/sfda-1E21/}{\textit{link}}.
\end{abstract}

%% file: 01intro.tex
\section{Introduction}
Reliable perception is a cornerstone of modern on-vehicle intelligent systems, supporting safety-critical applications such as autonomous driving and advanced driver-assistance systems. Object detection is a fundamental component of this pipeline, enabling downstream tasks, including tracking, motion prediction, and planning. Therefore, detection failures can propagate to higher-level decision modules and pose severe safety risks~\cite{DBLP:conf/cvpr/LeeFP13,DBLP:conf/cvpr/MaquedaL0GS18}. Despite substantial progress under favorable conditions, reliable object detection under adverse weather conditions remains a persistent problem. Fog, rain, snow, and low-illumination scenes induce \emph{structured} visibility degradations---such as contrast attenuation, color shifts, and partial occlusions---that systematically alter scene characteristics. Thus, detectors trained on clear-weather data degrade sharply when deployed in these conditions, with failures most pronounced on small or low-contrast objects such as pedestrians, riders, poles, and bicycles.

Domain adaptation~\cite{DBLP:journals/pami/OzaSVP24} provides a principled way to mitigate such distribution shifts, yet many real-world deployments cannot retain or access the original source training data. Privacy constraints, licensing restrictions, and storage or transmission costs often render the source images unavailable once a model is deployed. This motivates \emph{source-free domain adaptation} (SFDA)~\cite{DBLP:conf/iccv/YangW0HJ21}, where adaptation must proceed using only a source-trained detector and unlabeled target-domain images. Compared to standard unsupervised domain adaptation (UDA)~\cite{DBLP:conf/iccv/KodirovXFG15}, SFDA excludes explicit source–target alignment and makes adaptation highly
dependent on pseudo-label quality, often leading to unstable training. Taking this into account and benefiting from advanced diffusion models~\cite{DBLP:journals/corr/abs-2404-07771}, we incorporate layout-preserving weather variants synthesized using a frozen generative model, FLUX.1 KONTEXT~\cite{DBLP:journals/corr/abs-2506-15742}, to improve the  target-level structural consistency in various weather conditions.

Adverse weather conditions further amplify two failure modes that destabilize SFDA for detection. First, confidence-based filtering tends to \emph{over-prune} true positives: Many valid objects receive low confidence under visibility corruption and are discarded before contributing to learning. Second, existing SFDA methods largely rely on \emph{pairwise graph-based or instance-level representation learning objectives}, such as consistency regularization or contrastive alignment~\cite{DBLP:conf/cvpr/VSO023,DBLP:conf/iclr/Hwang0SY24}. Under severe corruption or strong augmentations, these objectives are prone to \emph{agreement collapse}, forcing alignment between semantically mismatched samples and reinforcing confirmation bias. 

We argue that SFDA under adverse weather should exploit \emph{group-level} structure that remains informative even when individual appearances are unreliable. In driving scenes, object proposals exhibit stable higher-order regularities, including contextual co-occurrence within an image, shared weather-dependent statistics, and latent semantic neighborhoods formed by confident instances. \framework operationalizes this insight by reformulating SFDA as \emph{hypergraph-guided representation learning}, as illustrated in Fig.~\ref{fig:data_proportion}. Specifically, we construct a contrastive hypergraph over target-domain proposals, where hyperedges encode complementary high-order relations: (i) intra-image contexts preserving scene structures, (ii) weather-aware groupings aligning domain-specific statistics, (iii) prototype-based clusters anchoring semantics to achieve class-consistent clustering across diverse weather conditions, and (iv) uncertainty-aware connections linking low-confidence proposals to reliable neighbors. We then introduce a group-wise \emph{Hyper-InfoNCE} objective that propagates semantic information through hyperedges, enabling robust adaptation under severe visibility degradation.
\input{figures/exampleConcept_wrap}
Beyond the hypergraph formulation, \framework is implemented within a teacher-student adaptation framework, where a teacher detector provides pseudo labels to supervise the student model during source-free adaptation. On top of this backbone, \framework incorporates several stabilizers tailored to adverse-weather adaptation. We refine noisy teacher predictions through a semantic-level hypergraph regularization module, which reduces cross-class leakage and mitigates confirmation bias in pseudo labels. To further strengthen cross-condition consistency, we synthesize layout-preserving weather variants using a frozen diffusion model, FLUX.1 KONTEXT~\cite{DBLP:journals/corr/abs-2506-15742}, as a data augmentation strategy, avoiding generator training or access to labeled source data. Together, these components improve pseudo-label reliability and provide more robust supervision for learning under severe visibility degradation.

To verify our proposed framework, we conduct a comprehensive experiment on multiple adverse-weather benchmarking datasets. The experimental results demonstrate that \framework consistently improves detection performance under severe visibility degradation and achieves state-of-the-art results in SFDA settings.

In summary, our contributions are: (i) We propose \framework, a hypergraph-guided source-free domain adaptation framework for object detection under adverse weather. (ii) We construct a multi-relation contrastive hypergraph over target-domain object proposals and optimize it using a group-wise Hyper-InfoNCE objective, enabling semantic propagation through high-order relations and mitigating agreement collapse. (iii) We incorporate prototype-based semantic regularization and uncertainty-aware propagation to stabilize pseudo-label learning for low-contrast instances under severe visibility degradation. (iv) We demonstrate the effectiveness of \framework through extensive experiments on multiple adverse-weather benchmarks under SFDA settings.

%% file: figures/exampleConcept_wrap.tex
\begin{wrapfigure}{r}{0.65\linewidth}
    \centering
    \includegraphics[width=\linewidth]{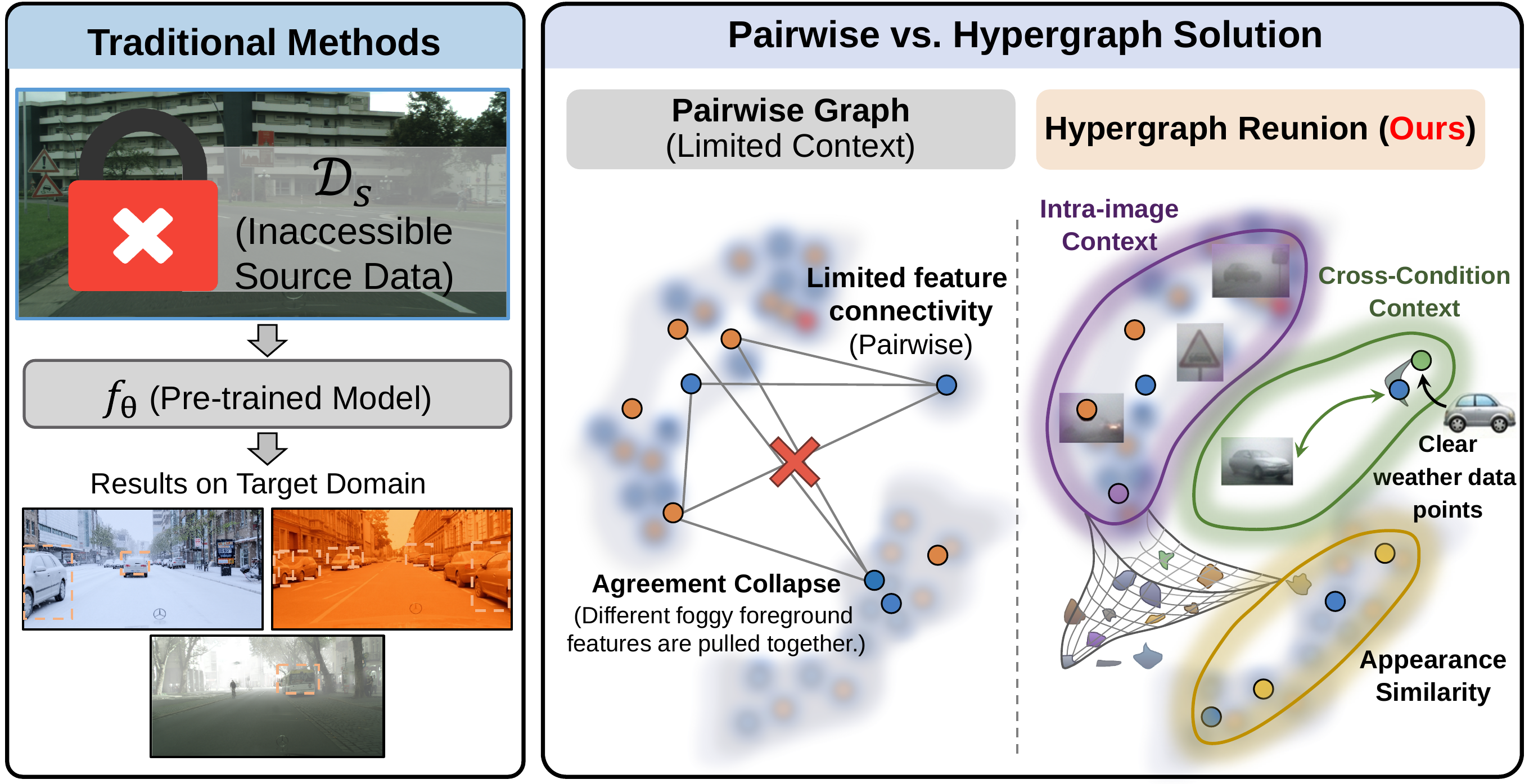}
    \captionsetup{font=small}
    \setlength{\abovecaptionskip}{1pt}
    \caption{\small In SFDA, source data is inaccessible, while target images suffer from structured degradation (e.g., fog, rain, night), leading to detection collapse. Existing pairwise contrastive methods fail in such noisy settings. \framework builds a contrastive hypergraph over object proposals using multi-relation hyperedges (e.g., intra-image context, and cross-condition views) to propagate semantics from confident to low-contrast instances.}
    \label{fig:data_proportion}
\end{wrapfigure}

%% file: 02related.tex
\section{Related Work}
\label{sec:relatedwork}
\subsection{Domain Adaptive Object Detection.}
Domain adaptive object detection (DAOD) tackles the domain gap between a labeled source domain and an unlabeled target domain by aligning representations or predictions across domains~\cite{DBLP:conf/nips/TarvainenV17}. Early and recent UDA-based DAOD methods adopt feature-level alignment, consistency regularization, and instance-aware objectives to mitigate distribution shift, yielding strong baselines such as MIC~\cite{DBLP:conf/cvpr/HoyerDWG23}, MRT~\cite{DBLP:conf/iccv/ZhaoWCLYPL23}, 2PC~\cite{DBLP:conf/cvpr/Deng0CD21}, AT~\cite{DBLP:conf/cvpr/LiDMLCWHKV22}, and ALDI~\cite{DBLP:journals/tmlr/KayHSDYPBH25}. These approaches typically assume access to source data during adaptation. Furthermore, source-free domain adaptation (SFDA) focuses on addressing the challenges posed by domain discrepancies solely with pseudo-labels generated by a source-pretrained model while source data is inaccessible. Existing thresholding-based methods~\cite{DBLP:conf/icml/LiangHF20,DBLP:conf/mm/YouLZCH21,DBLP:conf/cvpr/Li0ZZX22} rely solely on high-confidence predictions, discarding valuable information from uncertain candidates, yet potentially correct, detections. MixLRCo~\cite{DBLP:journals/corr/abs-2202-02802} propose low-confidence potential as pseudo-labels to maximize the utility of the target data. Beyond thresholding, other techniques have been proposed to refine pseudo-supervision. HCL~\cite{DBLP:conf/nips/HuangGXL21} employs contrastive learning for improved instance and category discrimination, while SOAP~\cite{DBLP:journals/ijis/XiongYZGLZ21} utilizes adversarial learning by injecting domain-specific noise into target images to regularize the detector. Upon this, to mitigate the domain shift, DiaNA~\cite{DBLP:conf/cvpr/HuangL0HCL23} and  DaC~\cite{DBLP:conf/nips/ZhangCCLLLL22} divide the target domain into source-similar and source-dissimilar subsets, employing adversarial learning to align the feature representations of both subsets, thereby enhancing the quality of the pseudo-supervision. Similarly, LODS~\cite{DBLP:conf/cvpr/Li0ZZX22} aims to improve robustness by learning representations that are invariant to domain-specific styles, directly tackling the negative effects of domain shift rather than just filtering its symptoms (i.e., noisy labels). Additionally, to conquer training instability and enhance model robustness, PETS~\cite{DBLP:conf/iccv/0004LSY23} mitigates it by introducing multiple teachers, fostering a more stable and robust adaptation process.
\subsection{Graph-based Approaches on Vision}
Graph-based approaches have been widely used in computer vision to model relational structure among visual entities such as regions or objects. Early works construct pairwise graphs to encode semantic and spatial relations for scene understanding and visual reasoning, most notably in scene graph generation and object-centric reasoning~\cite{DBLP:conf/cvpr/XuZCF17,DBLP:conf/eccv/LuKBL16,DBLP:conf/cvpr/HuGZDW18}. Such graph formulations have also been extended to video understanding by connecting spatial regions across time for spatio-temporal reasoning~\cite{DBLP:conf/eccv/WangG18}. However, these methods rely on pairwise edges and are limited in representing higher-order group structure in complex scenes. To overcome this limitation, recent works explore hypergraph formulations that model high-order relations via multi-node hyperedges. Representative approaches include Vision HGNN~\cite{DBLP:conf/iccv/HanW00W23} and HG-ViT~\cite{DBLP:conf/cvpr/Fixelle25}, which integrate hypergraph computation into recognition architectures, as well as hypergraph-based detection frameworks such as Hyper-YOLO~\cite{DBLP:journals/pami/FengHDYYLDJG25}. These methods demonstrate the effectiveness of hypergraphs for visual representation learning, but typically assume fully supervised training and use hypergraphs as forward computational modules. In contrast, our work leverages hypergraphs in an SFDA setting for object detection under adverse weather, constructing both proposal-level and semantic-level hypergraphs on unlabeled target data and using them to guide contrastive learning for robust semantic propagation.

%% file: 03metho.tex
\section{Methodology}
\label{sec:methodology}

\subsection{Problem Settings}
Let $\mathcal{X} = \{\mathbf{I} \in \mathbb{R}^{H \times W \times 3}\}$ denote the image space, where $\mathbf{I}$ represents an RGB image with height $H$ and width $W$.
Let $\mathcal{Y} = \{\mathbf{B}, \mathbf{c}, \mathbf{T}\}$ denote the detection output space, where $\mathbf{B}$ denotes bounding box coordinates, $\mathbf{c}$ denotes object category labels, and $\mathbf{T}$ denotes bounding box regression targets.

We consider a labeled source-domain detection dataset, $\mathcal{D}_{s} = \Big\{(\mathbf{I}^{s}_{i},\{\mathbf{B}^{s}_{i,r}, \mathbf{c}^{s}_{i,r},\mathbf{T}^{s}_{i,r}\}_{r})
\Big\}_{i=1}^{N_{s}}$,
where $\mathbf{I}^{s}_{i} \in \mathcal{X}$ is a source image and $\{\mathbf{B}^{s}_{i,r}, \mathbf{c}^{s}_{i,r}, \mathbf{T}^{s}_{i,r}\}_{r}$ denotes the set of annotated object instances in image $i$. 
Here, $\mathbf{B}^{s}_{i,r}$ is the bounding box of the $r$-th object, $\mathbf{c}^{s}_{i,r} \in \{1,\ldots,C\}$ is its class label, and $\mathbf{T}^{s}_{i,r}$ is the corresponding regression target.

A detector $F_{\theta_{0}}$, parameterized by $\theta_{0}$, is trained on $\mathcal{D}_{s}$ using a standard supervised detection objective:
\vspace{-2mm}
\begin{equation}
    \mathcal{L}^{s}_{\text{det}} = \frac{1}{N_s}\sum_{i=1}^{N_s}\sum_{r}\Big[\mathcal{L}_{\text{cls}}(f_{\theta_{0}}(\mathbf{B}^{s}_{i,r}), \mathbf{c}^{s}_{i,r}) + \lambda \mathcal{L}_{\text{reg}}(r_{\theta_{0}}(\mathbf{B}^{s}_{i,r}), \mathbf{T}^{s}_{i,r})\Big],
\end{equation}
\vspace{-1mm}
where $f_{\theta_{0}}$ and $r_{\theta_{0}}$ denote the classification and regression heads, respectively, and $\lambda$ balances the two loss terms.

After training, the source dataset $\mathcal{D}_{s}$ is no longer accessible due to privacy, storage, or transmission constraints.
Instead, we are given an unlabeled target-domain dataset $\mathcal{D}_{t} = \{ (\mathbf{I}^{t}_{i}, w_{i}) \}_{i=1}^{N_{t}}$, where $\mathbf{I}^{t}_{i}$ is a target image and $w_{i} \in \{1,\ldots,W\}$ is a discrete weather label (e.g., clear, rainy, foggy, snowy).
These weather labels are obtained from external sources such as metadata or a frozen, pretrained weather classifier~\cite{DBLP:conf/cvpr/HeZRS16}, and do not require access to source-domain data. 

Inspired by recent advances in the controllability of modern text-to-image (T2I) diffusion models, we use FLUX.1 KONTEXT~\cite{DBLP:journals/corr/abs-2506-15742} to synthesize layout-preserving cross-condition views for target images, producing semantically consistent variations across weather conditions while preserving scene structure. These generated images augment the target-domain set $\mathcal{D}_t$ and provide additional supervision for adaptation.

Due to adverse weather conditions, the source and target domains follow different joint distributions, i.e., $P_{s}(X,Y) \neq P_{t}(X,Y)$, where the shift is primarily caused by structured visibility degradation, including reduced contrast, color distortion, and occlusion~\cite{DBLP:conf/icra/JeonSM24}. Our objective is to adapt the pretrained detector from $\theta_{0}$ to new parameters $\theta$ using only unlabeled target data, such that the detection performance on the target domain is maximized.
\subsection{Framework Overview}

We adopt a mean-teacher framework~\cite{DBLP:conf/nips/TarvainenV17}, which consists of two detectors with identical architectures: a student model $F_{\theta}$ and a teacher model $F_{\phi}$.
The student model is optimized by gradient descent~\cite{DBLP:conf/cvpr/Deng0CD21}, while the teacher model is updated as an exponential moving average (EMA)~\cite{DBLP:conf/iclr/LaineA17} of the student via $\phi \leftarrow \alpha\phi + (1-\alpha)\theta$, 
where $\alpha \in (0,1)$ controls the temporal smoothness of the teacher. This formulation stabilizes training and provides more reliable pseudo labels for self-training. Since source data is inaccessible during adaptation, all region proposals, embeddings, and graph structures in the following are constructed exclusively from target-domain images; for notation simplicity, we omit the superscript $^{t}$ unless ambiguity may arise.

Given a target image $\mathbf{I}_{i}$, the detector generates a set of region proposals $\mathcal{B}_{i} = \{\mathbf{B}_{i,r}\}_{r=1}^{R_{i}}$ using a region proposal network (RPN)~\cite{DBLP:conf/nips/RenHGS15}. For each proposal $\mathbf{B}_{i,r}$, the teacher model outputs a class probability distribution $f_{\phi}(\mathbf{B}_{i,r})$ and a regression prediction $r_{\phi}(\mathbf{B}_{i,r})$.
\subsection{Pseudo Labeling}
Pseudo labels are constructed from the teacher predictions as $\hat{\mathbf{c}}_{i,r} = \arg\max_{c} f_{\phi}(\mathbf{B}_{i,r})_{c}, \quad \hat{\mathbf{T}}_{i,r} = r_{\phi}(\mathbf{B}_{i,r}).$ Using these pseudo labels, the student model is trained with an unsupervised detection loss:

\begin{equation}
    \begin{aligned}
        \mathcal{L}_{\text{det}} = \frac{1}{N_t} \sum_{i=1}^{N_t}\sum_{r}\Big[\mathcal{L}_{\text{cls}}(f_{\theta}(\mathbf{B}_{i,r}), \hat{\mathbf{c}}_{i,r}) + \lambda \mathcal{L}_{\text{reg}}(r_{\theta}(\mathbf{B}_{i,r}), \hat{\mathbf{T}}_{i,r})\Big].
    \end{aligned}
\end{equation}
To further enforce consistency between the student and teacher predictions, we introduce a knowledge distillation loss on the classification logits:
\begin{equation}
\mathcal{L}_{\text{KD}} = \frac{1}{N_t}
\sum_{i=1}^{N_t}\sum_{r} D_{\mathrm{KL}}\!\left[\sigma\!\left(\frac{f_{\phi}(\mathbf{B}_{i,r})}{\tau_{\text{KD}}}\right)\;\middle\|\;\sigma\!\left(\frac{f_{\theta}(\mathbf{B}_{i,r})}{\tau_{\text{KD}}}\right)\right],
\end{equation}
where $D_{\mathrm{KL}}(\cdot)$ denotes the Kullback-Leibler divergence, $\sigma(\cdot)$ denotes the \textit{softmax} function, and $f_{\phi}(\mathbf{B}_{i,r})$ and $f_{\theta}(\mathbf{B}_{i,r})$ represent the output classification logits of the teacher and student models, respectively. $\tau_{\text{KD}}$ is a temperature hyperparameter.
\subsection{Proposal-Level Hypergraph}
To capture higher-order structural relations among target proposals, we model proposals within a mini-batch as a weighted hypergraph.
Specifically, we flatten all region proposals in a mini-batch and index them using a single proposal index $p \in \{1,\ldots,N_b\}$, where $N_b$ denotes the total number of proposals in the mini-batch.
Each proposal $\mathbf{B}_{p}$ is treated as a vertex $v_{p}$.
We extract proposal-level embeddings from the teacher model and normalize them via $\displaystyle\mathbf{z}_{i} = \frac{g^{\text{emb}}_{\phi}(\mathbf{B}_{p})}   {\|g^{\text{emb}}_{\phi}(\mathbf{B}_{p})\|_{2}},$:
where $g^{\text{emb}}_{\phi}(\cdot)$ denotes the embedding head.

We construct three types of hyperedges.
First, feature-based hyperedges $\mathcal{E}_{\text{fea}}$ connect each proposal to its $K$ nearest neighbors in the embedding space, capturing local semantic similarity.
Second, intra-image hyperedges $\mathcal{E}_{\text{img}}$ connect all proposals originating from the same target image, encoding scene-level spatial and contextual coherence.
Third, weather-aware hyperedges $\mathcal{E}_{\text{wea}}$ connect proposals extracted from images sharing the same weather label, encouraging weather-invariant representation learning.

The full hyperedge set is $\mathcal{E} = \mathcal{E}_{\text{fea}} \cup \mathcal{E}_{\text{img}} \cup \mathcal{E}_{\text{wea}}$.
Let $\mathbf{H} \in \{0,1\}^{N_b \times |\mathcal{E}|}$ denote the hypergraph incidence matrix, where $\mathbf{H}_{p,e}=1$ indicates that vertex $v_{p}$ belongs to hyperedge $e$.
The induced adjacency matrix is defined as $\mathbf{A}_{p,q} = \sum_{e \in \mathcal{E}} w_{e}\mathbf{H}_{p,e}\mathbf{H}_{q,e}$
where $w_{e}$ denotes the weight of hyperedge $e$.
\subsection{Semantic-Level Hypergraph}
To provide global semantic anchors and mitigate pseudo-label drift, we maintain class- and weather-conditioned prototypes in the embedding space.
For each class $c \in \{1,\ldots,C\}$ and weather type $w \in \{1,\ldots,W\}$, we maintain a prototype vector $\mu_{c,w}$.
At each iteration, the prototype is updated using an EMA, $\displaystyle\mu_{c,w} \leftarrow (1-\gamma)\mu_{c,w} + \gamma \frac{1}{|\mathcal{S}_{c,w}|} \sum_{p \in \mathcal{S}_{c,w}} \mathbf{z}_{p},$
where $\mathcal{S}_{c,w}$ denotes the set of high-confidence proposals in the mini-batch assigned to class $c$ under weather $w$.
Proposals that are close to the same prototype $\mu_{c,w}$ in embedding space are grouped to form prototype-based hyperedges denoted as $\mathcal{E}_{\text{proto}}$, which encourage class-consistent clustering across weather conditions. Specifically, we define that proposal $v_p$ belongs to $\mathcal{E}_{\text{proto}}$ if its cosine similarity with the prototype exceeds a predefined threshold $\tau_{\text{proto}}$, as $\displaystyle\frac{\mathbf{z}_p^\top \mu_{c,w}}{\|\mathbf{z}_p\|_2 \|\mu_{c,w}\|_2} \ge \tau_{\text{proto}}$.

Adverse weather often causes true object instances to receive low-confidence predictions.
Instead of discarding these proposals, we explicitly model their relationships with reliable ones using uncertainty-aware hyperedges.

Teacher uncertainty is quantified using normalized entropy, $\displaystyle\mathcal{H}_{p} = -\frac{1}{\log C} \sum_{c} f_{\phi}(\mathbf{B}_{p})_{c} \log f_{\phi}(\mathbf{B}_{p})_{c}.$ Low-confidence proposals are connected to their $K$ nearest high-confidence neighbors, which is set by default at ten, that share the same pseudo class and weather condition. These connections form uncertainty-aware hyperedges denoted as $\mathcal{E}_{\text{unc}}$, whose weights are set proportional to $\mathcal{H}_{p}$. This design enables semantic information to propagate from reliable proposals to uncertain ones, improving robustness under severe visibility degradation. In addition, with the constructed hypergraph, let $Z \in \mathbb{R}^{N_b \times d}$ denote the proposal embedding matrix within a mini-batch, where the $p$-th row corresponds to $z_p$. Using the induced adjacency matrix $A$ composed of all hyperedges ($\mathcal{E}_{fea}, \mathcal{E}_{img}, \mathcal{E}_{wea}, \mathcal{E}_{proto}, \mathcal{E}_{unc}$), we perform a graph convolution step to aggregate features from structured neighbors, as $Z^{(l+1)} = \sigma(\tilde{D}^{-1/2} \tilde{A} \tilde{D}^{-1/2} Z^{(l)} W^{(l)})$, where $\tilde{A} = A + I$ is the adjacency matrix with self-loops, $\tilde{D}$ is the corresponding degree matrix ($\tilde{D}_{p,p} = \sum_q \tilde{A}_{p,q}$), $W^{(l)}$ is a learnable weight matrix. Through this forward message passing, isolated or low-confidence proposals dynamically absorb contextual and prototype-based semantics from reliable neighbors. The updated representations $Z^{(l+1)}$ are then utilized for the final detection predictions and the contrastive learning objective. In this way, the constructed HGNN acts as a relational regularizer before computing the contrastive loss.
\subsection{Hypergraph-Guided InfoNCE Learning Objective}
While the unsupervised detection and consistency losses enable self-training, they rely on individual pseudo labels and thus remain sensitive to noise introduced by adverse weather.
To further exploit the structural relations encoded in the proposal-level hypergraph, we introduce a hypergraph-guided contrastive learning objective that enforces representation consistency at the group level.

For each proposal $\mathbf{B}_{p}$, we extract $\ell_2$-normalized embeddings from both the student and teacher models:
\(  
    \displaystyle
    \mathbf{z}^{S}_{p} = \frac{g^{\text{emb}}_{\theta}(\mathbf{B}_{p})} {\|g^{\text{emb}}_{\theta}(\mathbf{B}_{p})\|_{2}}, \quad \mathbf{z}^{T}_{p} = \frac{g^{\text{emb}}_{\phi}(\mathbf{B}_{p})} {\|g^{\text{emb}}_{\phi}(\mathbf{B}_{p})\|_{2}},
\)
where $g^{\text{emb}}_{\theta}$ and $g^{\text{emb}}_{\phi}$ denote the embedding heads of the student and teacher, respectively.
We measure similarity using temperature-scaled cosine similarity via $\displaystyle\text{sim}(\mathbf{z}^{S}_{p}, \mathbf{z}^{T}_{q}) = \frac{(\mathbf{z}^{S}_{p})^{\top}\mathbf{z}^{T}_{q}}{\tau},$
where $\tau$ is a temperature hyperparameter.

For a given anchor proposal $p$, we define its positive set based on the hypergraph adjacency matrix $\mathbf{A}$.
Specifically, proposal $q$ is considered a positive of $p$ if (i) $p \neq q$, (ii) $\mathbf{A}_{p, q} > 0$, and (iii) both proposals share the same pseudo class label. Formally, the positive set is defined as $\mathcal{P}_{p} = \{ q \mid q \neq p,\; \mathbf{A}_{p, q} > 0,\; \hat{\mathbf{c}}_{q} = \hat{\mathbf{c}}_{p} \}$
This definition ensures that only structurally related and semantically compatible proposals are pulled together in the embedding space.

Different hyperedges encode relations of varying reliability.
To account for this, we weight each positive pair $(p, q)$ by the normalized hypergraph adjacency:
\(
\displaystyle
\tilde{w}_{p, q} =
\frac{\mathbf{A}_{p, q}}
{\sum_{k \in \mathcal{P}_{p}} \mathbf{A}_{p, k} + \epsilon},
\)
where $\epsilon$ is a small constant for numerical stability.

The hypergraph-guided InfoNCE loss for anchor $p$ is then defined as
\begin{equation}
\mathcal{L}_{\text{HG-NCE}} = - \sum_{q \in \mathcal{P}_{p}}
\tilde{w}_{p, q} \log \frac{\exp(\text{sim}(\mathbf{z}^{S}_{p}, \mathbf{z}^{T}_{q}))}{\sum_{k \neq p}\exp(\text{sim}(\mathbf{z}^{S}_{p}, \mathbf{z}^{T}_{k}))}.
\end{equation}
The denominator includes all proposals in the same mini-batch,
encouraging the student embedding to be close to hypergraph-consistent
positives while being separated from unrelated proposals.
\paragraph{Overall Hypergraph Contrastive Loss.}
The final hypergraph-guided InfoNCE objective is obtained by averaging over all anchors with non-empty positive sets:
\begin{equation}
\mathcal{L}_{\text{HG}} =
\frac{1}{|\mathcal{I}|}
\sum_{p \in \mathcal{I}}
\mathcal{L}_{\text{HG-NCE}}(p),
\quad
\mathcal{I} = \{ p \mid |\mathcal{P}_{p}| > 0 \}.
\end{equation}
This objective enforces consistency between student and teacher representations
at the level of hypergraph-defined groups, allowing semantic information to
propagate across intra-image context, weather conditions, prototype anchors,
and uncertainty-aware relations.
As a result, low-confidence proposals can recover meaningful representations
by leveraging their connections to reliable neighbors, which is critical under
severe visual degradation.

%% file: 04exper.tex
\section{Experiments}
\label{sec:experiments}

\subsection{Experimental Settings}
\label{subsec:experimental_settings}
\paragraph{Datasets and Senarios.}
The experiments are conducted on three datasets: Cityscapes~\cite{DBLP:journals/corr/CordtsORREBFRS16}, Cityscapes-Adverse, and Sim10k~\cite{DBLP:journals/corr/Johnson-Roberson16}. We conduct experiments on two scenarios: (1) City scene source-free adaptation: Cityscapes $\rightarrow$ Cityscapes-Adverse; (2) Sim2Real source-free adaptation: Sim10k $\rightarrow$ Cityscapes-Adverse.

\paragraph{Baselines.}
We compare \framework against representative object detection methods under domain shift, covering source-only, fully supervised, UDA, and SFDA settings. The Source-Only model, trained only on Cityscapes and directly evaluated on Cityscapes-Adverse, serves as the lower bound, while the Oracle model, trained with labeled target-domain data, provides an upper-bound reference. We also include several UDA methods, including MIC~\cite{DBLP:conf/cvpr/HoyerDWG23}, MRT~\cite{DBLP:conf/iccv/ZhaoWCLYPL23}, 2PC~\cite{DBLP:conf/cvpr/Deng0CD21}, AT~\cite{DBLP:conf/cvpr/LiDMLCWHKV22}, and ALDI~\cite{DBLP:journals/tmlr/KayHSDYPBH25}. These methods assume access to both source images and unlabeled target images during adaptation, and therefore serve as reference points rather than direct competitors. Our primary comparisons are with SFDA methods, including LPLD~\cite{DBLP:conf/eccv/YoonKKPJS24}, MT~\cite{DBLP:conf/nips/TarvainenV17}, DRU~\cite{DBLP:conf/eccv/KhanhNPTJ24}, IRG~\cite{DBLP:conf/cvpr/VSO023}, and SF-UT~\cite{DBLP:conf/eccv/HaoFF24}, which follow the same source-free constraint as \framework by adapting a source-pretrained detector using only unlabeled target-domain data.

\input{tables/sota}
\paragraph{Metrics.}
We evaluate object detection performance using three widely adopted metrics: Intersection over Union (IoU) and Average Precision (AP). We report AP computed at an IoU threshold of 0.5, following standard evaluation protocols in object detection.
\paragraph{Implementation and Hyperparameter Details.}

We adopt R-CNN with pre-trained ResNet-50 as the primary object detector. During the learning phase, the student network is initiated with a learning rate of 0.04, and we set the EMA for updating the teacher model to 0.9996. We set $\lambda = 1.0$ (see Equations 1 and 2) by default. All the experiments are conducted on four Nvidia A40 GPUs, with each GPU handling a batch size of 16. The implementation is carried out using the PyTorch framework and Detectron2~\cite{wu2019detectron2}.

\subsection{Experimental Findings}

As shown in Table~\ref{tab:sota}, we evaluate \framework in the setting Cityscapes $\rightarrow$ Cityscapes-Adverse against leading methods in both source-free domain adaptation (SFDA) and unsupervised domain adaptation (UDA). The proposed method achieves the best overall performance, improving AP by approximately 7.0\% over the strongest SFDA baseline, IRG~\cite{DBLP:conf/cvpr/VSO023}, while remaining competitive with several UDA approaches that have access to source data. 
Notably, \framework performs particularly well under challenging conditions such as rainy, sandstorm, and night-foggy. Qualitative results in Fig.~\ref{fig:city_comparsion} further illustrate that \framework produces more accurate detections across diverse illumination and weather conditions, including low-visibility scenes with artificial lighting. \input{tables/table2_wrap} In contrast, competing methods such as IRG~\cite{DBLP:conf/cvpr/VSO023} and SF-UT~\cite{DBLP:conf/eccv/HaoFF24} often produce false positives in dark regions or miss objects under severe illumination changes.
These results indicate that the proposed hypergraph-guided learning framework improves representation robustness across adverse weather conditions and varying object scales.
We further evaluate \framework in a challenging Sim2Real setting by adapting a detector trained on the synthetic Sim10k dataset to the real-world Cityscapes-Adverse benchmark. As shown in Table~\ref{tab:SHIFT}, \framework achieves the best performance among all SFDA methods, improving AP by 1.8\% over the strongest baseline, IRG~\cite{DBLP:conf/cvpr/VSO023}. Notably, our method also performs competitively with UDA approaches that rely on source data during adaptation.
This setting introduces substantial domain discrepancies, including differences in rendering realism, scene composition, and weather-induced visibility degradation. Despite these challenges, \framework consistently improves detection accuracy, demonstrating that hypergraph-guided representation learning effectively facilitates semantic transfer across both synthetic-to-real and adverse-weather domain shifts.
\input{figures/vis}
We further compare \framework with SFDA baselines on Cityscapes $\rightarrow$ Cityscapes-Adverse scenarios according to different APs for different object sizes. As shown in Table~\ref{tab:object_detection1}, \framework achieves the best performance on large and small objects, reaching 34.6 AP\textsuperscript{L} and 8.4 AP\textsuperscript{S}. These gains indicate that the proposed hypergraph formulation effectively propagates semantic information across related proposals, helping maintain reliable representations despite weather-induced degradation. Improvements are particularly evident for large objects, where contextual cues are more informative. While gains on small objects remain limited due to extremely low visibility, the results demonstrate that hypergraph-guided learning improves detection robustness under adverse conditions.
\subsection{Ablation Study}

To assess the contribution of each component in \framework, we performed ablation experiments on the Cityscapes $\rightarrow$ Cityscapes-Adverse benchmark in different configurations. \input{tables/tab3_wrap} As shown in Table~\ref{tab:ablation_main}, the baseline model using only weak-strong augmentation achieves limited performance. Introducing pairwise augmentation improves the performance to 34.1\% AP. However, relying solely on pairwise contrastive views is insufficient to bridge the large domain gap caused by structured visibility degradation, often leading to agreement collapse. 
By incorporating the proposal-level and semantic-level hyperedges, \framework achieves 38.7\% AP even without the Hyper-InfoNCE objective, indicating that modeling higher-order relations among proposals provides more reliable semantic cues for learning under visual corruption. This is attributed to the forward message passing mechanism defined in REUNION. Specifically, the HGNN serves as a structural prior and provides relational regularization that guides the student model's representation learning, allowing confident instances to calibrate the features of corrupted ones (e.g., via uncertainty-aware edges) before they are passed to the detection head. Thus, the HGNN prevents the student model from collapsing under severe weather degradation by achieving relational consistency at the feature level.

\input{tables/ablation_combined}
To further analyze the contribution of different hyperedge types, we perform incremental experiments as shown in Table~\ref{tab:ablation_edges}. Starting with feature-based hyperedges ($\mathcal{E}_{\text{fea}}$), the model captures local semantic similarity among proposals. Adding intra-image ($\mathcal{E}_{\text{img}}$) and weather-aware ($\mathcal{E}_{\text{wea}}$) hyperedges improves performance by 0.7\% AP, demonstrating the benefit of incorporating scene-level context and domain-specific grouping. 
Further introducing prototype-based ($\mathcal{E}_{\text{proto}}$) and uncertainty-aware ($\mathcal{E}_{\text{unc}}$) hyperedges yields the best performance of 39.3\% AP on Cityscapes-Adverse. These hyperedges anchor proposal features to stable teacher prototypes and connect low-confidence proposals to reliable neighbors, enabling the model to recover informative instances that would otherwise be discarded under severe visibility degradation. Overall, integrating all hyperedge families enables \framework to better bridge the domain gap between clear and adverse weather conditions.
\input{figures/figVis1}
In addition, Fig.~\ref{fig:class-wiseAP} shows the class-wise AP across different weather conditions. \framework consistently outperforms existing SFDA methods across most categories, with larger gains under challenging conditions such as rain, snow, sandstorm, and night-foggy. This improvement highlights the benefit of modeling higher-order relations among proposals. Unlike pairwise methods, the proposed hypergraph enables semantic propagation among related instances, allowing the model to recover reliable representations for corrupted or low-confidence proposals and maintain stable detection performance across diverse weather conditions.

%% file: tables/sota.tex
\begin{table*}[htb] 
\centering
\setlength{\tabcolsep}{2pt} 
\caption{Performance comparison on ResNet50 (in AP) of our method with other approaches for UDA/SFDA from Cityscapes to Cityscapes-Adverse.}
\label{tab:sota}
\scalebox{0.85}{
\begin{tabular}{@{}c|c| ccccccc|c@{}}
\hline
\toprule
Type & Method & D-Rainy & D-Snowy & D-Foggy & D-Sandstorm & N-Clear & N-Rainy & N-Foggy   & mAP \\ \hline
Lower-Bound &   Source-Only                       & 15.9      & 16.1      &  20.8     &  3.0        &  9.2        & 6.4         & 5.5         &       11.0               \\ \hline
Higher-Bound            & Oracle      & 41.9                          & 50.9     & 48.1      & 45.7      & 39.7         & 35.5         &  37.1        &  42.7                           \\ \hline
\multirow{5}{*}{UDA}    
& MIC~\cite{DBLP:conf/cvpr/HoyerDWG23}                         &21.0       &23.2       & 29.9      & 26.2         & 17.3         & 10.9         & 13.8         & 20.3                \\ 

& MRT~\cite{DBLP:conf/iccv/ZhaoWCLYPL23}                         &18.4       &25.2       &29.6      & 18.3         & 17.3         & 7.0         & 14.6         & 18.6           \\
& 2PC~\cite{DBLP:conf/cvpr/Deng0CD21}                         &37.4       &31.3       & 42.5      & 17.0         & 27.9         & 28.6         & 22.1         & 29.5                \\
& AT~\cite{DBLP:conf/cvpr/LiDMLCWHKV22}                         &32.1       &26.4       &41.7      &35.5         &27.6         & 27.4         & 19.7         & 30.1                 \\
& ALDI~\cite{DBLP:journals/tmlr/KayHSDYPBH25}                          & 39.8       &36.0       & 54.1      & 44.0         & 35.5         & 27.9         & 26.6         & 37.7               \\                        \hline
\multirow{6}{*}{SFDA}
&  LPLD~\cite{DBLP:conf/eccv/YoonKKPJS24}                        & 25.4      & 25.3      & 34.9      & 30.7         & 22.7         & 13.9         & 19.3         &    24.6                  \\ 
&  MT~\cite{DBLP:conf/nips/TarvainenV17}                         & 26.1      & 26.7      & 37.8      & 26.9         & 24.1         & 12.9         & 19.0         &    24.8                  \\ 
& DRU~\cite{DBLP:conf/eccv/KhanhNPTJ24}                        & 32.1      & 30.6      & 42.4      & 32.3         & 27.1         & 16.5         & 20.7         &    28.8                  \\            
&  SF-UT~\cite{DBLP:conf/eccv/HaoFF24}                        & 30.8      & \underline{33.1}      & \underline{48.0}      & \underline{37.9}         & 29.0         & 14.5         & 23.6         & 31.0                     \\
&  IRG~\cite{DBLP:conf/cvpr/VSO023}                     & \underline{35.0}      & 33.0      & 33.8      & 34.4         &  \underline{30.4}        & \textbf{31.8}         &  \underline{27.6}         & \underline{32.3}                     \\
\hline
\rowcolor{gray!10} & \framework (\textbf{ours})                 & \textbf{43.7}      & \textbf{38.2}      & \textbf{53.9}      & \textbf{45.7}         & \textbf{36.1}         & \underline{26.6}         & \textbf{30.8}        & \textbf{39.3}                   \\  
\bottomrule
\end{tabular}
}
\end{table*}

%% file: tables/table2_wrap.tex
\begin{wraptable}{r}{0.26\linewidth}
    \centering
    \captionsetup{font=small}
    \setlength{\tabcolsep}{1.2pt}
    \caption{Sim2Real Adaptation from Sim10K $\rightarrow$ Cityscapes-Adverse.}
    \label{tab:SHIFT}

    \renewcommand{\arraystretch}{0.68}
    \scriptsize
    \resizebox{\linewidth}{!}{ 
        \begin{tabular}{@{}l|l|c@{}}
            \toprule[1.0pt]
            Type & Method & AP Car \\ 
            \midrule
            S & Source Only & 19.5 \\ 
            \midrule
            \multirow{7}{*}{UDA} 
             & DAF~\cite{DBLP:conf/cvpr/Chen0SDG18} & 22.4 \\
             & PT~\cite{DBLP:conf/icml/ChenCYSWZYQZXP22} & 22.7 \\
             & MIC~\cite{DBLP:conf/cvpr/HoyerDWG23} & 27.5 \\
             & AT~\cite{DBLP:conf/cvpr/LiDMLCWHKV22} & 28.4 \\ 
             & 2PC~\cite{DBLP:conf/cvpr/Deng0CD21} & 31.2 \\
             & ALDI~\cite{DBLP:journals/tmlr/KayHSDYPBH25} & 42.4 \\
             \midrule
            \multirow{6}{*}{SFDA} 
             & LPLD~\cite{DBLP:conf/eccv/YoonKKPJS24} & 32.0 \\
             & MT~\cite{DBLP:conf/nips/TarvainenV17} & 33.1 \\
             & DRU~\cite{DBLP:conf/eccv/KhanhNPTJ24} & 37.0 \\
             & SF-UT~\cite{DBLP:conf/eccv/HaoFF24} & 39.7 \\
             & IRG~\cite{DBLP:conf/cvpr/VSO023} & 40.8 \\
             & \framework (\textbf{ours}) & 42.6 \\ 
             \bottomrule[1.0pt]
        \end{tabular}
    }
\end{wraptable}

%% file: figures/vis.tex
\begin{figure*}[htbp]
\begin{center}
\includegraphics[width=0.80\linewidth]{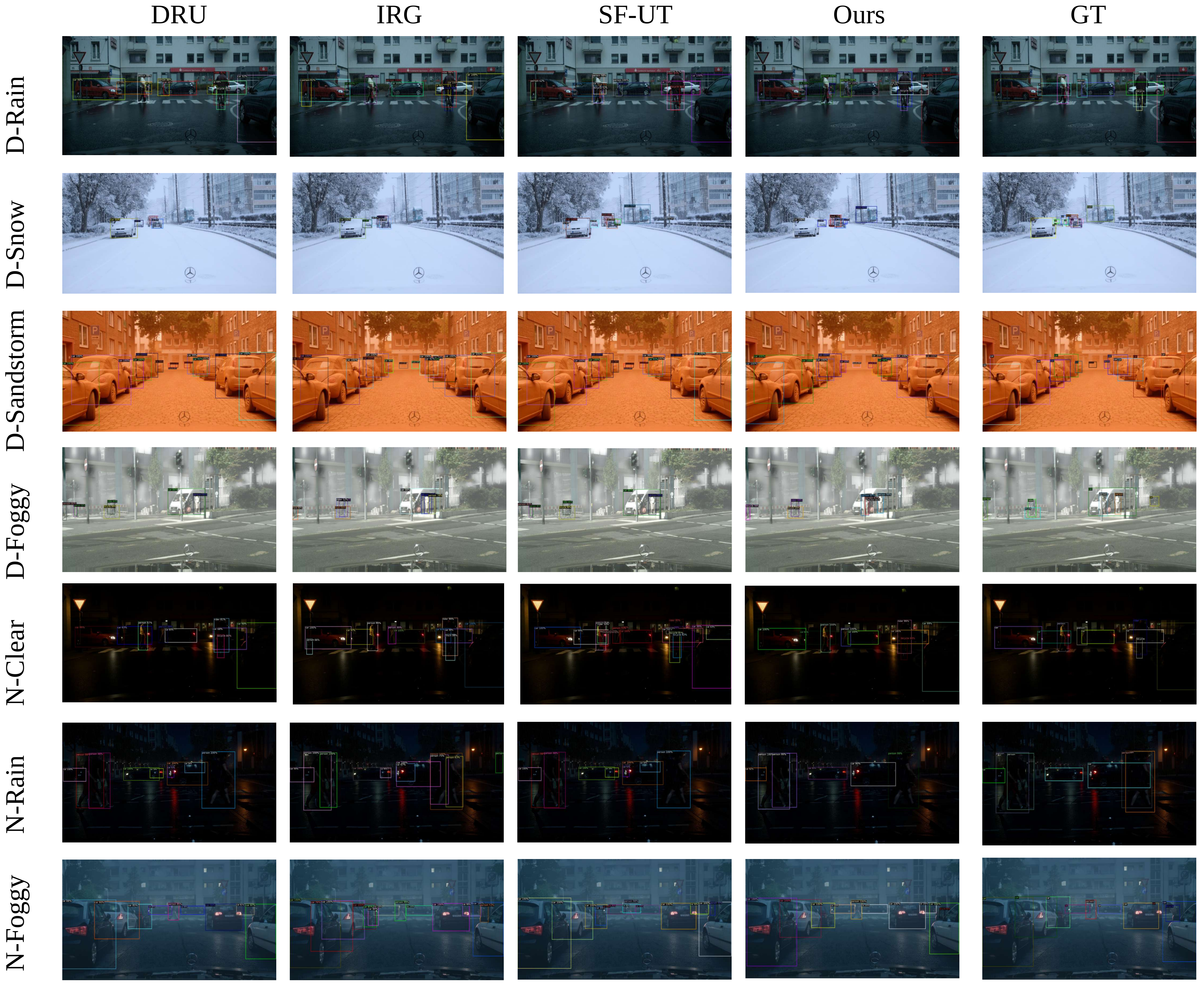}
\end{center}
\setlength{\abovecaptionskip}{1pt}
\caption{
Qualitative results for the Cityscapes-Adverse dataset are displayed from left to right, showcasing the raw image, followed by results from DRU~\cite{DBLP:conf/eccv/KhanhNPTJ24}, IRG~\cite{DBLP:conf/cvpr/VSO023}, SF-UT~\cite{DBLP:conf/eccv/HaoFF24}, and the proposed, with the ground truth presented on the far right. 
}
\label{fig:city_comparsion}
\end{figure*}

%% file: tables/tab3_wrap.tex
\begin{wraptable}{r}{0.35\linewidth}
    \centering
    \captionsetup{font=small}
    \setlength{\tabcolsep}{1.2pt}
    \caption{Comparison to SFDA baselines with different object sizes on Cityscapes-Adverse.}
    \label{tab:object_detection1}

    \renewcommand{\arraystretch}{1.05}
    \resizebox{\linewidth}{!}{
        \begin{tabular}{@{}l|l|ccc@{}}
        \toprule[1.0pt]
        Type & Method & AP\textsuperscript{S} & AP\textsuperscript{M} & AP\textsuperscript{L} \\
        \midrule
        \multirow{5}{*}{SFDA} 
         & LPLD~\cite{DBLP:conf/eccv/YoonKKPJS24} & 5.2 & 14.3 & 19.4 \\
         & MT~\cite{DBLP:conf/nips/TarvainenV17} & 5.3 & 15.6 & 19.8 \\
         & IRG~\cite{DBLP:conf/cvpr/VSO023} & 7.4 & 12.4 & 29.4 \\
         & SF-UT~\cite{DBLP:conf/eccv/HaoFF24} & 6.7 & 11.8 & 28.0 \\
         & \framework (\textbf{ours}) & 8.4 & 11.1 & 34.6 \\
        \bottomrule[1.0pt]
        \end{tabular}
    }
\end{wraptable}

%% file: tables/ablation_combined.tex
\begin{table}[!htbp]
    \centering
    \captionsetup{font=small}
    \begin{minipage}[t]{0.55\linewidth}
        \centering
        \caption{Ablation study on Cityscapes-Adverse.}
        \label{tab:ablation_main}
        \resizebox{\linewidth}{!}{
            \begin{tabular}{c|c|c|c|cccc}
            \toprule[1.5pt]
            \multirow{2}{*}{PairAug} & 
            \multirow{2}{*}{\begin{tabular}[c]{@{}c@{}}Proposal\\ HGNN\end{tabular}} & \multirow{2}{*}{\begin{tabular}[c]{@{}c@{}}Semantic\\ HGNN\end{tabular}} & \multirow{2}{*}{\begin{tabular}[c]{@{}c@{}}Hyper-\\Info-NCE\end{tabular}} & \multirow{2}{*}{AP} & 
            \multirow{2}{*}{AP\textsuperscript{L}} & 
            \multirow{2}{*}{AP\textsuperscript{S}} & 
            \multirow{2}{*}{AP\textsuperscript{M}} \\ 
             & & & & & & & \\ \hline
            & & & & 30.0 & 28.6 & 6.9 & 10.7 \\ \hline
            \checkmark & & & & 34.1 & 30.7 & 7.5 & 10.8 \\
            \checkmark & & \checkmark & & 37.1 & 33.0 & 7.9 & 11.0 \\
            \checkmark & \checkmark & & & 37.9 & 33.7 & 8.1 & 11.6 \\
            \checkmark & \checkmark & \checkmark & & 38.7 & 34.1 & 8.2 & 11.5  \\
            \hline
            \checkmark & \checkmark & \checkmark & \checkmark & 39.3 & 34.6 & 8.4 & 11.1 \\ 
            \bottomrule[1.5pt]
            \end{tabular}
        }
    \end{minipage}
    \hfill %
    \begin{minipage}[t]{0.44\linewidth}
        \centering
        \caption{Impact of hyperedge types in Hypergraph-based SFDA.}
        \label{tab:ablation_edges}
        \resizebox{\linewidth}{!}{
            \begin{tabular}{ccccc|cccc}
            \toprule[1.5pt]
            \multicolumn{5}{c|}{Hyperedge Types ($\mathcal{E}^{(t)}$)} & \multirow{2}{*}{AP} & \multirow{2}{*}{AP\textsuperscript{L}} & \multirow{2}{*}{AP\textsuperscript{S}} & \multirow{2}{*}{AP\textsuperscript{M}} \\
             Fea & Img & Weath & Proto & Unc & & & & \\ \midrule
             \checkmark & & & & & 37.2 & 33.0 & 7.9 & 11.1 \\
             \checkmark & \checkmark & & & & 37.4 & 33.2 & 8.1 & 11.6 \\
             \checkmark & \checkmark & \checkmark & & & 37.9 & 33.7 & 8.1 & 11.6 \\
             \checkmark & \checkmark & \checkmark & \checkmark & & 38.8 & 33.7 & 7.9 & 9.3 \\ \hline
             \checkmark & \checkmark & \checkmark & \checkmark & \checkmark & 39.3 & 34.6 & 8.4 & 11.1 \\ \bottomrule[1.5pt]
            \end{tabular}
        }
    \end{minipage}
\end{table}

%% file: figures/figVis1.tex

\begin{figure*}[!htbp]
    \centering
    \includegraphics[width=\linewidth]{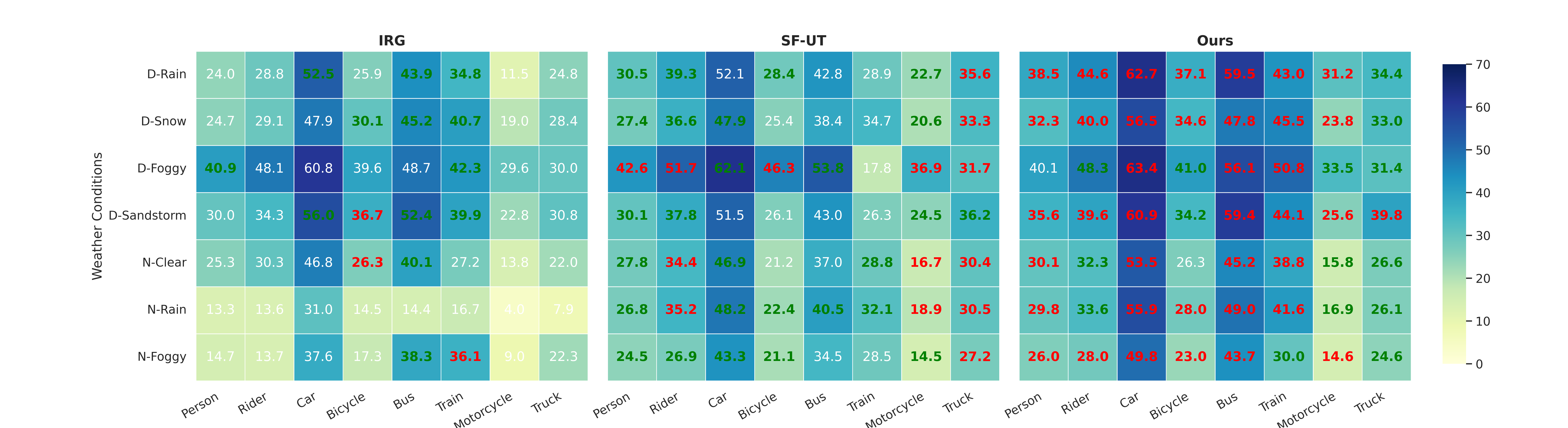}
    \captionsetup{font=small}
    \setlength{\abovecaptionskip}{1pt}
    \setlength{\belowcaptionskip}{0pt}
    \caption{\textbf{Class-wise AP across weather conditions.}
    Comparison of IRG~\cite{DBLP:conf/cvpr/VSO023}, SF-UT~\cite{DBLP:conf/eccv/HaoFF24}, and \framework; best and second-best results are highlighted in \textcolor{red}{red} and \textcolor{green}{green}, respectively.}
    \label{fig:class-wiseAP}
\end{figure*}

%% file: 05conclu.tex
\section{Conclusion and Future Work~\label{FW}}
We presented \framework, a source-free domain adaptation framework for object detection under adverse weather. \framework reformulates SFDA as hypergraph-guided representation learning, organizing proposal relations with intra-image context, weather-aware grouping, prototype anchors, and uncertainty-aware connections. The group-wise Hyper-InfoNCE objective propagates semantics from reliable to low-confidence instances and mitigates agreement collapse under severe visibility degradation. Experiments on Cityscapes-Adverse and cross-domain settings demonstrate consistent gains over strong UDA and SFDA baselines, validating the value of higher-order relational constraints in source-free adaptation. Future work includes extending \framework to broader adverse conditions (mixed weather and low-light), improving pseudo-label calibration under severe corruption, and developing lighter hypergraph message-passing schemes for better efficiency.

%% file: 07appendix.tex
\appendix
\section{Supplementary Material}

The supplementary material provides additional details to complement the main paper. In particular, we offer further insights into the experimental datasets used in our study and present additional ablation experiments to analyze the contribution of different components of \framework. These results provide a deeper understanding of the proposed method and further support its effectiveness for source-free domain adaptation in adverse-weather object detection.

\subsection{Datasets and Evaluation Metrics}
\label{app:dataset}
We evaluate the proposed \framework on the Cityscapes-Adverse benchmark for source-free domain adaptation (SFDA) in adverse-weather object detection. Our experiments involve both real-world and synthetic datasets, including Cityscapes~\cite{DBLP:journals/corr/CordtsORREBFRS16}, Sim10K~\cite{DBLP:journals/corr/Johnson-Roberson16}, and Cityscapes-Adverse. The Cityscapes-Adverse dataset is constructed from Cityscapes by introducing seven types of adverse weather and lighting conditions. In total, it contains 23,600 images covering eight object categories. The datasets are divided into training, adaptation, and evaluation splits following standard protocols~\cite{DBLP:conf/cvpr/VSO023,DBLP:conf/iclr/Hwang0SY24}.

\input{tables/data_stat}

For the Cityscapes $\rightarrow$ Cityscapes-Adverse scenario, we train the source detector using 2,950 daytime images from Cityscapes. During adaptation, 20,650 unlabeled images from Cityscapes-Adverse are used, while evaluation is performed on 3,500 Cityscapes-Adverse images. 

For the Sim10K $\rightarrow$ Cityscapes-Adverse scenario, the source model is trained using 9,975 images from Sim10K. During adaptation, 19,768 unlabeled images from Cityscapes-Adverse are used, and the evaluation set contains 3,346 images from the same dataset.

Importantly, no target-domain annotations are used during the adaptation stage in either scenario. Table~\ref{tab:data_stat} summarizes the statistics of the experimental datasets.

Next, we analyze the distribution of object classes in Cityscapes, as illustrated in Fig.~\ref{fig:dataStat}. The class distribution in Cityscapes-Adverse follows a similar pattern, since each Cityscapes image is used to generate seven corresponding images under different light and adverse-weather conditions. In contrast, the Sim10K dataset contains only a single object category, namely \textit{car}.

\input{figures/figDataStat}

\subsection{Baseline Details}
\label{app:baselines}

To comprehensively evaluate the performance of \framework, we compare it with a wide range of state-of-the-art object detection methods under domain shift. The baselines are grouped according to their adaptation settings and data requirements, as summarized in Table~\ref{tab:sota}.

The Source-Only model is trained exclusively on the source domain, Cityscapes, and evaluated directly on the target domain, Cityscapes-Adverse, without any adaptation. This baseline quantifies the performance degradation caused by the domain shift and serves as a lower-bound reference. In contrast, the Oracle model is trained using labeled target-domain data and represents an upper-bound reference for adaptation performance.

We also compare with several unsupervised domain adaptation (UDA) methods, including MIC~\cite{DBLP:conf/cvpr/HoyerDWG23}, MRT~\cite{DBLP:conf/iccv/ZhaoWCLYPL23}, 2PC~\cite{DBLP:conf/cvpr/Deng0CD21}, AT~\cite{DBLP:conf/cvpr/LiDMLCWHKV22}, and ALDI~\cite{DBLP:journals/tmlr/KayHSDYPBH25}. These methods operate under a more privileged setting than ours, since they require access to both source-domain images and unlabeled target-domain images during adaptation. Therefore, they are not direct competitors to our source-free setting, but they provide useful reference points for understanding the performance achievable when source data is available for alignment.

The most relevant comparisons are with source-free domain adaptation (SFDA) methods, including LPLD~\cite{DBLP:conf/eccv/YoonKKPJS24}, MT~\cite{DBLP:conf/nips/TarvainenV17}, DRU~\cite{DBLP:conf/eccv/KhanhNPTJ24}, IRG~\cite{DBLP:conf/cvpr/VSO023}, and SF-UT~\cite{DBLP:conf/eccv/HaoFF24}. These methods follow the same challenging setting as \framework: they adapt a source-pretrained detector using only unlabeled target-domain data, without access to the original source images. Thus, they serve as the direct competitors for evaluating the effectiveness of our proposed approach in the practical source-free setting.

\subsection{Additional Experiments}
Table~\ref{tab:sota_full} presents the detailed experimental results. The observations are consistent with those reported in the main paper. In particular, the proposed method achieves the best overall performance, improving AP by approximately 7.0\% over the strongest SFDA baseline, IRG~\cite{DBLP:conf/cvpr/VSO023}, while remaining competitive with several UDA approaches that have access to source data. Notably, \framework performs especially well under challenging conditions such as rainy, sandstorm, and night-foggy.

\input{tables/sota_full}

\subsection{Additional Ablation Study}
\subsubsection{Effect of Confidence Neighbors \texorpdfstring{$K$}{K}.}

To analyze the impact of the uncertainty-aware hyperedges ($\mathcal{E}_{\text{unc}}$), we evaluate the model under different values of $K$, which controls the number of high-confidence neighbors connected to each low-confidence proposal. As shown in Table~\ref{tab:ablationK}, varying $K$ influences how effectively semantic information propagates from reliable anchors to uncertain instances under severe visibility degradation.

\input{tables/ablationK}

When $K$ is too small (e.g., $K=6$ or $K=8$), low-confidence proposals receive limited contextual support, making the learned representations sensitive to noisy predictions and local feature variations. As a result, detection performance across object scales (AP\textsuperscript{L}, AP\textsuperscript{M}, and AP\textsuperscript{S}) remains suboptimal. 

Increasing $K$ to 10 yields the best performance, as a larger set of reliable neighbors forms a more stable semantic neighborhood that helps recover fragmented features caused by adverse weather and challenging light conditions. However, further increasing $K$ (e.g., $K=12$ or $K=14$) slightly degrades performance, since overly large neighborhoods may introduce semantically mismatched connections and lead to negative transfer. Therefore, $K=10$ provides the best trade-off between semantic richness and noise robustness.

\subsubsection{Effect of Prototype Updating
\texorpdfstring{$\gamma$}{gamma}.}
We maintain class- and weather-conditioned prototypes to provide global semantic anchors and mitigate pseudo-label drift. Table~\ref{tab:ablationGamma} reports the impact of the prototype updating momentum parameter $\gamma$, which balances the contribution of the current mini-batch against the historical moving average (see Equation~9 in the main paper). 

The results show that the adaptation process is sensitive to the choice of $\gamma$. A high value (e.g., $\gamma=0.99$) causes the prototypes to update too aggressively from the current mini-batch. Under severe visibility degradation, this leads to unstable training because noisy pseudo-labels and weather-induced artifacts can drift the semantic anchors. In contrast, an excessively low value (e.g., $\gamma=0.85$) makes the prototypes overly rigid, preserving historical representations and limiting the model's ability to adapt to the target weather and light conditions.

The best performance is achieved at intermediate values (e.g., $\gamma=0.90$ or $0.95$), where the prototypes stabilize training while remaining flexible enough to capture evolving weather- and light-invariant semantics. This balanced update particularly benefits large (AP\textsuperscript{L}) and medium (AP\textsuperscript{M}) objects, where consistent global semantic clustering is important for reliable domain adaptation.

\subsection{Performance Comparison to the state of the art VLMs/MLMs}
Table~\ref{tab:object_detection} compares \framework with existing SFDA vision specialists as well as recent Vision-Language Models (VLMs) and Multimodal Foundation Models (MFMs). Among the vision specialists, \framework achieves the best performance on both small and large objects, reaching 8.4 AP\textsuperscript{S} and 34.6 AP\textsuperscript{L}. The improvement on large objects is particularly significant, indicating that the proposed hypergraph formulation effectively propagates semantic information across related proposals and stabilizes pseudo-label learning under severe visibility degradation.

In contrast, general-purpose VLMs and MFMs exhibit substantially lower detection performance in this setting. While some models (e.g., Gemini 2.5 Flash and Qwen 3.0-VL) achieve good performance on large objects, their results on small and medium objects remain very limited. This gap highlights the difficulty of applying general multimodal models directly to dense detection tasks under adverse weather without task-specific adaptation. Overall, these results demonstrate that specialized SFDA methods remain more effective for domain-adaptive object detection, while \framework further advances the state of the art by leveraging hypergraph-guided relational learning.

\input{tables/vsLM}

Fig.~\ref{fig:city_comparsion_supp} presents additional qualitative comparisons between \framework and several VLM/MFM baselines under diverse adverse-weather and light conditions. Consistent with the quantitative results in Table~\ref{tab:object_detection}, \framework produces more accurate and stable detections in challenging scenarios such as heavy rain, snow, fog, and nighttime environments. In particular, our method is able to correctly localize objects that are partially occluded or severely degraded by weather artifacts, whereas competing approaches often miss these instances or generate false positives in low-contrast regions.

Among the VLMs/MFMs, Qwen 2.5-VL largely fails to detect objects in the adverse-weather dataset. The newer Qwen 3.0-VL shows significant improvement and achieves performance close to \framework. Similarly, Gemini~2.5 Flash demonstrates relatively strong detection capability and produces results comparable to our method in several cases. In contrast, GPT-4.1 fails to produce meaningful detection results. Although GPT-4.1 can often infer the presence of objects in the scene, its bounding-box localization is extremely inaccurate, resulting in nearly zero overall AP. This limitation has also been reported in prior studies~\cite{DBLP:journals/inffus/SapkotaK26,DBLP:journals/corr/abs-2410-19808}.

The visualization also reveals remaining failure cases in extremely challenging scenarios, such as nighttime rain, where strong reflections and severe illumination shifts continue to degrade detection reliability. Nevertheless, the overall qualitative results further demonstrate that \framework improves the robustness and consistency of object detection across diverse adverse-weather conditions.

%% file: tables/data_stat.tex
\begin{table}[h]
\centering
\small
\caption{Dataset statistics.}
\label{tab:data_stat}
\begin{tabular}{lcccc}
\toprule[0.8pt]
\multirow{2}{*}{\textbf{Dataset}} & \multirow{2}{*}{\textbf{No. Classes}} & \multicolumn{3}{c}{\textbf{No. Images}} \\ \cline{3-5} 
& & Source &  Adaptation & Evaluation \\ \hline \hline
Cityscapes\cite{DBLP:journals/corr/CordtsORREBFRS16} & 8         & 2,950       & -            &     -       \\
Sim10K\cite{DBLP:journals/corr/Johnson-Roberson16} & 1         & 9,975        & -           &     -     \\
\hline
Cityscapes $\rightarrow$ Cityscapes-Adverse & 8         &     -  &     20,650        & 3,500            \\
Sim10K $\rightarrow$ Cityscapes-Adverse & 1         &    -    &     19,768       & 3,346           \\
\bottomrule[0.8pt]
\end{tabular}
\end{table}

%% file: figures/figDataStat.tex
\begin{figure*}[h]
\begin{center}
\includegraphics[width=0.95\linewidth]{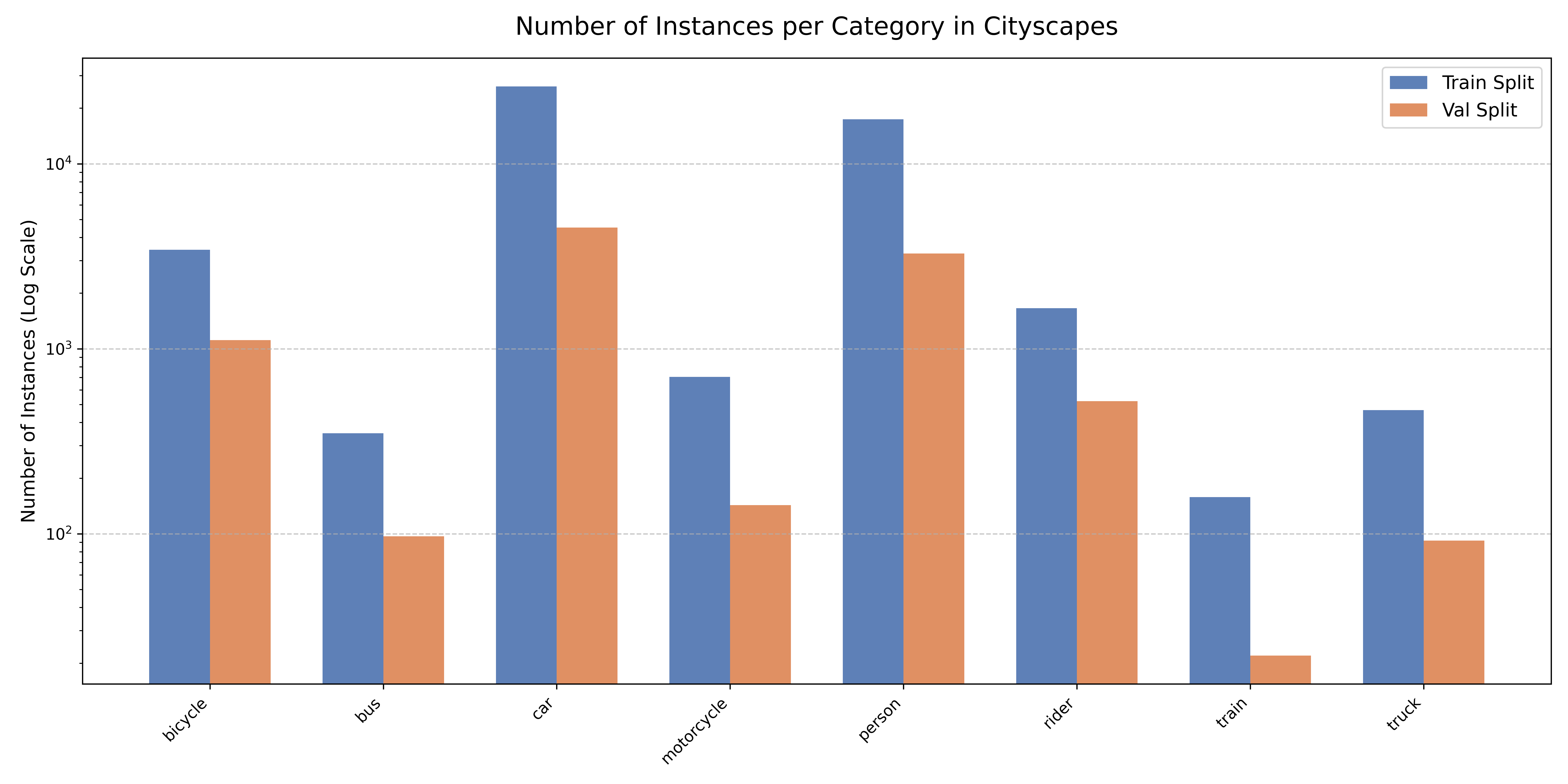}
\end{center}
\setlength{\abovecaptionskip}{1pt}
\captionsetup{font=small}
\caption{Class-wise instances distribution from Cityscapes dataset.}
\label{fig:dataStat}
\end{figure*}

%% file: tables/sota_full.tex
\begin{table*}[htb] 
\caption{Performance comparison on ResNet50 (in AP) of our method with other approaches for UDA/SFDA from Cityscapes to Cityscapes-Adverse.}
\label{tab:sota_full}
\centering
\scalebox{0.8}{

\begin{tabular}{@{}c|c| ccccccc|c@{}}
\hline
\toprule
Type & Methods & D-Rainy & D-Snowy & D-Foggy & D-Sandstorm & N-Clear & N-Rainy & N-Foggy   & mAP \\ \hline
Lower-Bound &   Source-Only                       & 15.9      & 16.1      &  20.8     &  3.0        &  9.2        & 6.4         & 5.5         &       11.0               \\ \hline
Higher-Bound            & Oracle      & 41.9                          & 50.9     & 48.1      & 45.7      & 39.7         & 35.5         &  37.1        &  42.7                           \\ \hline
\multirow{3}{*}{UDA}    
& DAF~\cite{DBLP:conf/cvpr/Chen0SDG18}                         &25.0       &23.2       & 32.8      & 30.0         & 18.6         & 11.1         & 15.3         & 22.3 \\
& SADAF~\cite{DBLP:journals/ijcv/ChenWLSDG21}                         &21.6       &22.9       & 33.7      & 27.4         & 17.0         & 9.0         & 15.2         & 20.1 \\
& PT~\cite{DBLP:conf/icml/ChenCYSWZYQZXP22}                        &23.6       &21.5       & 35.9      & 28.3         & 18.6         & 16.0         & 14.7         & 22.7 \\
& MIC~\cite{DBLP:conf/cvpr/HoyerDWG23}                         &21.0       &23.2       & 29.9      & 26.2         & 17.3         & 10.9         & 13.8         & 20.3                \\ 

& MRT~\cite{DBLP:conf/iccv/ZhaoWCLYPL23}                       &18.4       &25.2       &29.6      & 18.3         & 17.3         & 7.0         & 14.6         & 18.6           \\
& 2PC~\cite{DBLP:conf/cvpr/Deng0CD21}                         &37.4       &31.3       & 42.5      & 17.0         & 27.9         & 28.6         & 22.1         & 29.5                \\
& AT~\cite{DBLP:conf/cvpr/LiDMLCWHKV22}                         &32.1       &26.4       &41.7      &35.5         &27.6         & 27.4         & 19.7         & 30.1                 \\
& ALDI~\cite{DBLP:journals/tmlr/KayHSDYPBH25}                          & 39.8       &36.0       & 54.1      & 44.0         & 35.5         & 27.9         & 26.6         & 37.7               \\                        \hline
\multirow{3}{*}{SFDA}
& PETS~\cite{DBLP:conf/iccv/0004LSY23}                    & 15.8      & 16.6      & 24.6      & 15.4         & 16.9         & 10.3         & 12.1         &    16.0                  \\
&  LPLD~\cite{DBLP:conf/eccv/YoonKKPJS24}                        & 25.4      & 25.3      & 34.9      & 30.7         & 22.7         & 13.9         & 19.3         &    24.6                  \\ 
&  MT~\cite{DBLP:conf/nips/TarvainenV17}                         & 26.1      & 26.7      & 37.8      & 26.9         & 24.1         & 12.9         & 19.0         &    24.8                  \\ 
& DRU~\cite{DBLP:conf/eccv/KhanhNPTJ24}                        & 32.1      & 30.6      & 42.4      & 32.3         & 27.1         & 16.5         & 20.7         &    28.8                  \\            
&  SF-UT~\cite{DBLP:conf/eccv/HaoFF24}                        & 30.8      & \underline{33.1}      & \underline{48.0}      & \underline{37.9}         & 29.0         & 14.5         & 23.6         & 31.0                     \\
&  IRG~\cite{DBLP:conf/cvpr/VSO023}                     & \underline{35.0}      & 33.0      & 33.8      & 34.4         &  \underline{30.4}        & \textbf{31.8}         &  \underline{27.6}         & \underline{32.3}                     \\
\hline   
\rowcolor{gray!10} 
& \framework                  & \textbf{43.7}      & \textbf{38.2}      & \textbf{53.9}      & \textbf{45.7}         & \textbf{36.1}         & \underline{26.6}         & \textbf{30.8}        & \textbf{39.3}                   \\                     
\bottomrule
\end{tabular}
}
\end{table*}

%% file: tables/ablationK.tex
\begin{table}[ht]
    \centering
    \begin{minipage}[t]{0.48\textwidth}
        \centering
        \caption{Effect of Confidence Neighbors $K$.}
        \label{tab:ablationK}
        \scalebox{0.9}{
        \begin{tabular}{c| c c c}
        \toprule
        $K$ & AP\textsuperscript{S} & AP\textsuperscript{M} & AP\textsuperscript{L} \\
        \midrule
        6 & 7.9 & 10.4 & 33.1 \\
        8 & 8.2 & 10.8 & 34.0 \\
        10 & \textbf{8.4} & \textbf{11.1} & \textbf{34.6} \\
        12 & 8.1 & 10.8 & 33.7 \\
        14  & 7.8 & 10.1 & 32.7 \\
        \bottomrule
        \end{tabular}}
    \end{minipage}\hfill
    \begin{minipage}[t]{0.48\textwidth}
        \centering
        \caption{Effect of Prototype Updating $\gamma$.} %
        \label{tab:ablationGamma}
        \scalebox{0.9}{
        \begin{tabular}{c| c c c}
        \toprule
        \textbf{$\gamma$} & AP\textsuperscript{S} & AP\textsuperscript{M} & AP\textsuperscript{L} \\
        \midrule
        0.85 & 8.0 & 10.7 & 34.0 \\
        0.90 & \textbf{8.4} & 11.1 & \textbf{34.6} \\
        0.95 & 8.0 & \textbf{11.4} & 34.2 \\
        0.99 & 8.1 & 10.9 & 34.1 \\
        \bottomrule
        \end{tabular}}
    \end{minipage}
\end{table}

%% file: tables/vsLM.tex
\begin{table}[ht]
\caption{Quantitative Comparison of Object Detection Capabilities. The performance of our proposed method is compared against state-of-the-art Source-Free Domain Adaptation (SFDA) vision specialists and modern Vision-Language Models (VLMs) / Multi-modal Foundation Models (MFMs).}
\centering
\scalebox{0.9}{
\begin{tabular}{l l c c c}
\toprule
\textbf{Baselines} & \textbf{Model} & AP\textsuperscript{S} & AP\textsuperscript{M} & AP\textsuperscript{L} \\
\midrule
\multirow{5}{*}{Vision Specialists} 
& LPLD~\cite{DBLP:conf/eccv/YoonKKPJS24} & 5.2 & 14.3 & 19.4 \\
& MT~\cite{DBLP:conf/nips/TarvainenV17} & 5.3 & \textbf{15.6} & 19.8 \\
& IRG~\cite{DBLP:conf/cvpr/VSO023} & 7.4 & 12.4 & 29.4 \\
& SF-UT~\cite{DBLP:conf/eccv/HaoFF24} & 6.7 & 11.8 & 28.0 \\
\rowcolor{gray!10} 
& \framework (\textbf{ours}) & \textbf{8.4} & 11.1 & \textbf{34.6} \\
\midrule
\multirow{7}{*}{VLMs / MFMs}
& Grounding Dino~\cite{DBLP:journals/corr/abs-2410-19808} & 0.0 & 2.5 & 19.2 \\
 & GPT-4o~\cite{DBLP:journals/corr/abs-2410-21276} & 0.0 & 0.0 & 0.0 \\
 & GPT-4.1~\cite{DBLP:journals/corr/abs-2303-08774} & 0.0 & 0.0 & 0.5 \\
 & Gemini 2.5 Flash~\cite{DBLP:journals/corr/abs-2507-06261} & 0.1 & 4.9 & 29.9 \\

 & Qwen 2.5-VL~\cite{DBLP:journals/corr/abs-2502-13923} & 0.0 & 0.0 & 0.0 \\
 & Qwen 3.0-VL~\cite{DBLP:journals/corr/abs-2511-21631} & 0.1 & 5.4 & 31.6 \\
  & Google OWLv2~\cite{DBLP:conf/nips/MindererGH23} & 0.0 & 3.5 & 21.9 \\
\bottomrule
\end{tabular}}
\label{tab:object_detection}
\end{table}
\input{figures/supplement-vis}

%% file: figures/supplement-vis.tex
\begin{figure*}[htbp]
    \begin{subfigure}[t]{1\textwidth}
      \centering
      \includegraphics[width=1\textwidth]{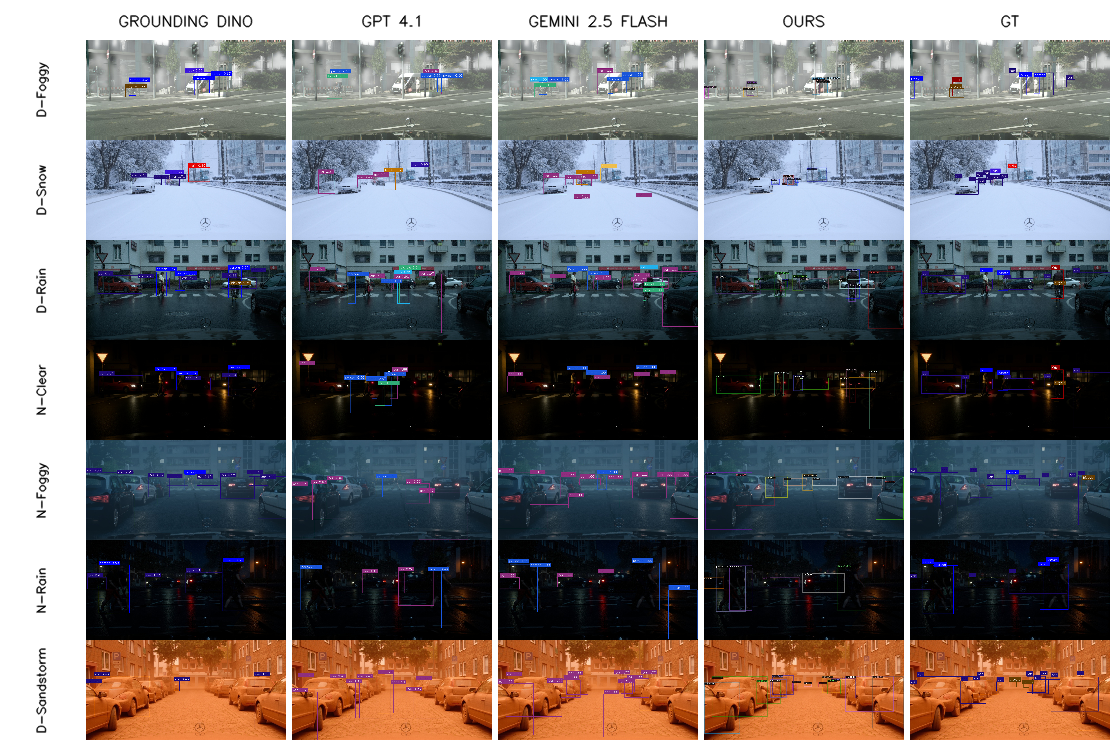}
    \end{subfigure}
    \begin{subfigure}[t]{1\textwidth}
      \centering
      \includegraphics[width=1\textwidth]{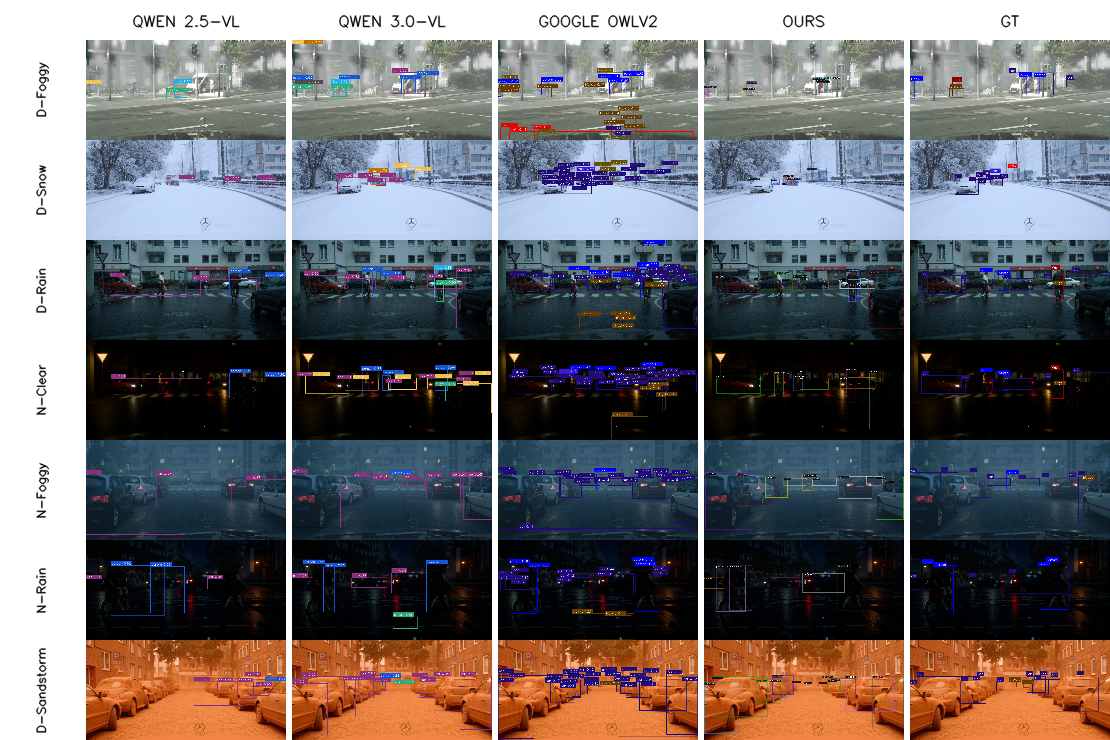}
    \end{subfigure}
\caption{
Qualitative results for the Cityscapes-Adverse dataset are displayed from left to right, showcasing the raw image, followed by results from Grounding Dino, GPT-4.1, Gemini 2.5 Flash, Qwen 2.5-VL, Qwen 3.0-VL, Google OWLv2 and the proposed (Ours), with the ground truth (GT) presented on the far right. 
}
\label{fig:city_comparsion_supp}
\end{figure*}

%% file: main.bbl
\begin{thebibliography}{25}
\providecommand{\natexlab}[1]{#1}
\providecommand{\EM}{\em}
\providecommand{\RNtxt}{\relax}
\RNtxt{}

\bibitem[Bai et~al.(2025)S.~Bai, K.~Chen, X.~Liu, J.~Wang, W.~Ge, S.~Song, K.~Dang, P.~Wang, S.~Wang, J.~Tang, H.~Zhong, Y.~Zhu, M.~Yang, Z.~Li, J.~Wan, P.~Wang, W.~Ding, Z.~Fu, Y.~Xu, J.~Ye, X.~Zhang, T.~Xie, Z.~Cheng, H.~Zhang, Z.~Yang, H.~Xu, J.~Lin]{DBLP:journals/corr/abs-2502-13923}
{\EM Bai Shuai, Chen Keqin, Liu Xuejing, Wang Jialin, Ge~Wenbin, Song Sibo, Dang Kai, Wang Peng, Wang Shijie, Tang Jun, Zhong Humen, Zhu Yuanzhi, Yang Ming{-}Hsuan, Li~Zhaohai, Wan Jianqiang, Wang Pengfei, Ding Wei, Fu~Zheren, Xu~Yiheng, Ye~Jiabo, Zhang Xi, Xie Tianbao, Cheng Zesen, Zhang Hang, Yang Zhibo, Xu~Haiyang, Lin Junyang}.
\newblock Qwen2.5-VL Technical Report \allowbreak\newblock// CoRR. 2025. abs/2502.13923.

\bibitem[Chen et~al.(2022)M.~Chen, W.~Chen, S.~Yang, J.~Song, X.~Wang, L.~Zhang, Y.~Yan, D.~Qi, Y.~Zhuang, D.~Xie, S.~Pu]{DBLP:conf/icml/ChenCYSWZYQZXP22}
{\EM Chen Meilin, Chen Weijie, Yang Shicai, Song Jie, Wang Xinchao, Zhang Lei, Yan Yunfeng, Qi~Donglian, Zhuang Yueting, Xie Di, Pu~Shiliang}.
\newblock Learning Domain Adaptive Object Detection with Probabilistic Teacher \allowbreak\newblock// Proceedings of the International Conference on Machine Learning (ICML). 2022.  3040--3055.

\bibitem[Chen et~al.(2018)Y.~Chen, W.~Li, C.~Sakaridis, D.~Dai, L.~V. Gool]{DBLP:conf/cvpr/Chen0SDG18}
{\EM Chen Yuhua, Li~Wen, Sakaridis Christos, Dai Dengxin, Gool Luc~Van}.
\newblock Domain Adaptive Faster {R-CNN} for Object Detection in the Wild \allowbreak\newblock// Proceedings of the {IEEE/CVF} Conference on Computer Vision and Pattern Recognition (CVPR). 2018.  3339--3348.

\bibitem[Chen et~al.(2021)Y.~Chen, H.~Wang, W.~Li, C.~Sakaridis, D.~Dai, L.~V. Gool]{DBLP:journals/ijcv/ChenWLSDG21}
{\EM Chen Yuhua, Wang Haoran, Li~Wen, Sakaridis Christos, Dai Dengxin, Gool Luc~Van}.
\newblock Scale-Aware Domain Adaptive Faster {R-CNN} \allowbreak\newblock// Int. J. Comput. Vis. 2021. 129, 7. 2223--2243.

\bibitem[Chiang et~al.(2024)T.~Chiang, J.~Robinson, X.~V. Yu, D.~Yogatama]{DBLP:journals/corr/abs-2410-19808}
{\EM Chiang Ting{-}Rui, Robinson Joshua, Yu~Xinyan~Velocity, Yogatama Dani}.
\newblock LocateBench: Evaluating the Locating Ability of Vision Language Models \allowbreak\newblock// CoRR. 2024. abs/2410.19808.

\bibitem[Cordts et~al.(2016)M.~Cordts, M.~Omran, S.~Ramos, T.~Rehfeld, M.~Enzweiler, R.~Benenson, U.~Franke, S.~Roth, B.~Schiele]{DBLP:journals/corr/CordtsORREBFRS16}
{\EM Cordts Marius, Omran Mohamed, Ramos Sebastian, Rehfeld Timo, Enzweiler Markus, Benenson Rodrigo, Franke Uwe, Roth Stefan, Schiele Bernt}.
\newblock The Cityscapes Dataset for Semantic Urban Scene Understanding \allowbreak\newblock// CoRR. 2016. abs/1604.01685.

\bibitem[Deng et~al.(2021)J.~Deng, W.~Li, Y.~Chen, L.~Duan]{DBLP:conf/cvpr/Deng0CD21}
{\EM Deng Jinhong, Li~Wen, Chen Yuhua, Duan Lixin}.
\newblock Unbiased Mean Teacher for Cross-Domain Object Detection \allowbreak\newblock// Proceedings of the {IEEE/CVF} Conference on Computer Vision and Pattern Recognition (CVPR). 2021.  4091--4101.

\bibitem[Hao et~al.(2024)Y.~Hao, F.~Forest, O.~Fink]{DBLP:conf/eccv/HaoFF24}
{\EM Hao Yan, Forest Florent, Fink Olga}.
\newblock Simplifying Source-Free Domain Adaptation for Object Detection: Effective Self-training Strategies and Performance Insights \allowbreak\newblock// Proceedings of the European Conference on Computer Vision (ECCV). 2024.  196--213.

\bibitem[Hoyer et~al.(2023)L.~Hoyer, D.~Dai, H.~Wang, L.~V. Gool]{DBLP:conf/cvpr/HoyerDWG23}
{\EM Hoyer Lukas, Dai Dengxin, Wang Haoran, Gool Luc~Van}.
\newblock {MIC:} Masked Image Consistency for Context-Enhanced Domain Adaptation \allowbreak\newblock// Proceedings of the {IEEE/CVF} Conference on Computer Vision and Pattern Recognition (CVPR). 2023.  11721--11732.

\bibitem[Hwang et~al.(2024)U.~Hwang, J.~Lee, J.~Shin, S.~Yoon]{DBLP:conf/iclr/Hwang0SY24}
{\EM Hwang Uiwon, Lee Jonghyun, Shin Juhyeon, Yoon Sungroh}.
\newblock {SF(DA)2:} Source-free Domain Adaptation Through the Lens of Data Augmentation \allowbreak\newblock// Proceedings of the International Conference on Learning Representations (ICLR). 2024.

\bibitem[Johnson{-}Roberson et~al.(2016)M.~Johnson{-}Roberson, C.~Barto, R.~Mehta, S.~N. Sridhar, R.~Vasudevan]{DBLP:journals/corr/Johnson-Roberson16}
{\EM Johnson{-}Roberson Matthew, Barto Charles, Mehta Rounak, Sridhar Sharath~Nittur, Vasudevan Ram}.
\newblock Driving in the Matrix: Can Virtual Worlds Replace Human-Generated Annotations for Real World Tasks? \allowbreak\newblock// CoRR. 2016. abs/1610.01983.

\bibitem[Kay et~al.(2025)J.~Kay, T.~Haucke, S.~Stathatos, S.~Deng, E.~Young, P.~Perona, S.~Beery, G.~V. Horn]{DBLP:journals/tmlr/KayHSDYPBH25}
{\EM Kay Justin, Haucke Timm, Stathatos Suzanne, Deng Siqi, Young Erik, Perona Pietro, Beery Sara, Horn Grant~Van}.
\newblock Align and Distill: Unifying and Improving Domain Adaptive Object Detection \allowbreak\newblock// Trans. Mach. Learn. Res. 2025. 2025.

\bibitem[Khanh et~al.(2024)T.~L.~B. Khanh, H.~Nguyen, L.~H. Pham, D.~N. Tran, J.~W. Jeon]{DBLP:conf/eccv/KhanhNPTJ24}
{\EM Khanh Trinh Le~Ba, Nguyen Huy{-}Hung, Pham Long~Hoang, Tran Duong~Nguyen{-}Ngoc, Jeon Jae~Wook}.
\newblock Dynamic Retraining-Updating Mean Teacher for Source-Free Object Detection \allowbreak\newblock// Proceedings of the European Conference on Computer Vision (ECCV). 2024.  328--344.

\bibitem[Li et~al.(2022)Y.~Li, X.~Dai, C.~Ma, Y.~Liu, K.~Chen, B.~Wu, Z.~He, K.~Kitani, P.~Vajda]{DBLP:conf/cvpr/LiDMLCWHKV22}
{\EM Li~Yu{-}Jhe, Dai Xiaoliang, Ma~Chih{-}Yao, Liu Yen{-}Cheng, Chen Kan, Wu~Bichen, He~Zijian, Kitani Kris, Vajda Peter}.
\newblock Cross-Domain Adaptive Teacher for Object Detection \allowbreak\newblock// Proceedings of the {IEEE/CVF} Conference on Computer Vision and Pattern Recognition (CVPR). 2022.  7571--7580.

\bibitem[Liu et~al.(2023)Q.~Liu, L.~Lin, Z.~Shen, Z.~Yang]{DBLP:conf/iccv/0004LSY23}
{\EM Liu Qipeng, Lin Luojun, Shen Zhifeng, Yang Zhifeng}.
\newblock Periodically Exchange Teacher-Student for Source-Free Object Detection \allowbreak\newblock// Proceedings of the {IEEE/CVF} International Conference on Computer Vision (ICCV). 2023.  6391--6401.

\bibitem[Minderer et~al.(2023)M.~Minderer, A.~A. Gritsenko, N.~Houlsby]{DBLP:conf/nips/MindererGH23}
{\EM Minderer Matthias, Gritsenko Alexey~A., Houlsby Neil}.
\newblock Scaling Open-Vocabulary Object Detection \allowbreak\newblock// Proceedings of the Annual Conference on Neural Information Processing Systems (NeurIPS). 2023.

\bibitem[OpenAI(2023)]{DBLP:journals/corr/abs-2303-08774}
{\EM OpenAI }.
\newblock {GPT-4} Technical Report \allowbreak\newblock// CoRR. 2023. abs/2303.08774.

\bibitem[OpenAI(2024)]{DBLP:journals/corr/abs-2410-21276}
{\EM OpenAI }.
\newblock GPT-4o System Card \allowbreak\newblock// CoRR. 2024. abs/2410.21276.

\bibitem[Sapkota, Karkee(2026)R.~Sapkota, M.~Karkee]{DBLP:journals/inffus/SapkotaK26}
{\EM Sapkota Ranjan, Karkee Manoj}.
\newblock Object Detection with Multimodal Large Vision-Language Models: An In-depth Review \allowbreak\newblock// Inf. Fusion. 2026. 126. 103575.

\bibitem[Tarvainen, Valpola(2017)A.~Tarvainen, H.~Valpola]{DBLP:conf/nips/TarvainenV17}
{\EM Tarvainen Antti, Valpola Harri}.
\newblock Mean Teachers Are Better Role Models: Weight-averaged Consistency Targets Improve Semi-supervised Deep Learning Results \allowbreak\newblock// Proceedings of the Annual Conference on Neural Information Processing Systems (NeurIPS). 2017.  1195--1204.

\bibitem[Team(2025{\natexlab{a}})G.~Team]{DBLP:journals/corr/abs-2507-06261}
{\EM Team Gemini}.
\newblock Gemini 2.5: Pushing the Frontier with Advanced Reasoning, Multimodality, Long Context, and Next Generation Agentic Capabilities \allowbreak\newblock// CoRR. 2025{\natexlab{a}}. abs/2507.06261.

\bibitem[Team(2025{\natexlab{b}})Q.~Team]{DBLP:journals/corr/abs-2511-21631}
{\EM Team Qwen}.
\newblock Qwen3-VL Technical Report \allowbreak\newblock// CoRR. 2025{\natexlab{b}}. abs/2511.21631.

\bibitem[VS et~al.(2023)V.~VS, P.~Oza, V.~M. Patel]{DBLP:conf/cvpr/VSO023}
{\EM VS~Vibashan, Oza Poojan, Patel Vishal~M.}
\newblock Instance Relation Graph Guided Source-Free Domain Adaptive Object Detection \allowbreak\newblock// Proceedings of the {IEEE/CVF} Conference on Computer Vision and Pattern Recognition (CVPR). 2023.  3520--3530.

\bibitem[Yoon et~al.(2024)I.~Yoon, H.~Kwon, J.~Kim, J.~Park, H.~Jang, K.~Sohn]{DBLP:conf/eccv/YoonKKPJS24}
{\EM Yoon Ilhoon, Kwon Hyeongjun, Kim Jin, Park Junyoung, Jang Hyunsung, Sohn Kwanghoon}.
\newblock Enhancing Source-Free Domain Adaptive Object Detection with Low-Confidence Pseudo Label Distillation \allowbreak\newblock// Proceedings of the European Conference on Computer Vision (ECCV).  15142. 2024.  337--353.

\bibitem[Zhao et~al.(2023)Z.~Zhao, S.~Wei, Q.~Chen, D.~Li, Y.~Yang, Y.~Peng, Y.~Liu]{DBLP:conf/iccv/ZhaoWCLYPL23}
{\EM Zhao Zijing, Wei Sitong, Chen Qingchao, Li~Dehui, Yang Yifan, Peng Yuxin, Liu Yang}.
\newblock Masked Retraining Teacher-Student Framework for Domain Adaptive Object Detection \allowbreak\newblock// Proceedings of the {IEEE/CVF} International Conference on Computer Vision (ICCV). 2023.  18993--19003.

\end{thebibliography}


\begin{thebibliography}{10}

\bibitem{DBLP:conf/icml/ChenCYSWZYQZXP22}
Meilin Chen, Weijie Chen, Shicai Yang, Jie Song, Xinchao Wang, Lei Zhang, Yunfeng Yan, Donglian Qi, Yueting Zhuang, Di~Xie, and Shiliang Pu.
\newblock Learning domain adaptive object detection with probabilistic teacher.
\newblock In {\em Proceedings of the International Conference on Machine Learning (ICML)}, pages 3040--3055, 2022.

\bibitem{DBLP:journals/corr/abs-2404-07771}
Minshuo Chen, Song Mei, Jianqing Fan, and Mengdi Wang.
\newblock An overview of diffusion models: Applications, guided generation, statistical rates and optimization.
\newblock {\em CoRR}, abs/2404.07771, 2024.

\bibitem{DBLP:conf/cvpr/Chen0SDG18}
Yuhua Chen, Wen Li, Christos Sakaridis, Dengxin Dai, and Luc~Van Gool.
\newblock Domain adaptive faster {R-CNN} for object detection in the wild.
\newblock In {\em Proceedings of the {IEEE/CVF} Conference on Computer Vision and Pattern Recognition (CVPR)}, pages 3339--3348, 2018.

\bibitem{DBLP:journals/corr/CordtsORREBFRS16}
Marius Cordts, Mohamed Omran, Sebastian Ramos, Timo Rehfeld, Markus Enzweiler, Rodrigo Benenson, Uwe Franke, Stefan Roth, and Bernt Schiele.
\newblock The cityscapes dataset for semantic urban scene understanding.
\newblock {\em CoRR}, abs/1604.01685, 2016.

\bibitem{DBLP:conf/cvpr/Deng0CD21}
Jinhong Deng, Wen Li, Yuhua Chen, and Lixin Duan.
\newblock Unbiased mean teacher for cross-domain object detection.
\newblock In {\em Proceedings of the {IEEE/CVF} Conference on Computer Vision and Pattern Recognition (CVPR)}, pages 4091--4101, 2021.

\bibitem{DBLP:journals/pami/FengHDYYLDJG25}
Yifan Feng, Jiangang Huang, Shaoyi Du, Shihui Ying, Jun{-}Hai Yong, Yipeng Li, Guiguang Ding, Rongrong Ji, and Yue Gao.
\newblock Hyper-yolo: When visual object detection meets hypergraph computation.
\newblock {\em {IEEE} Trans. Pattern Anal. Mach. Intell.}, 47(4):2388--2401, 2025.

\bibitem{DBLP:conf/cvpr/Fixelle25}
Joshua Fixelle.
\newblock Hypergraph vision transformers: Images are more than nodes, more than edges.
\newblock In {\em Proceedings of the IEEE/CVF Conference on Computer Vision and Pattern Recognition (CVPR)}, pages 9751--9761, 2025.

\bibitem{DBLP:conf/iccv/HanW00W23}
Yan Han, Peihao Wang, Souvik Kundu, Ying Ding, and Zhangyang Wang.
\newblock Vision {HGNN:} an image is more than a graph of nodes.
\newblock In {\em Proceedings of the IEEE/CVF Conference on Computer Vision (ICCV)}, pages 19821--19831, 2023.

\bibitem{DBLP:conf/eccv/HaoFF24}
Yan Hao, Florent Forest, and Olga Fink.
\newblock Simplifying source-free domain adaptation for object detection: Effective self-training strategies and performance insights.
\newblock In {\em Proceedings of the European Conference on Computer Vision (ECCV)}, pages 196--213, 2024.

\bibitem{DBLP:conf/cvpr/HeZRS16}
Kaiming He, Xiangyu Zhang, Shaoqing Ren, and Jian Sun.
\newblock Deep residual learning for image recognition.
\newblock In {\em Proceedings of the {IEEE/CVF} Conference on Computer Vision and Pattern Recognition (CVPR)}, pages 770--778, 2016.

\bibitem{DBLP:conf/cvpr/HoyerDWG23}
Lukas Hoyer, Dengxin Dai, Haoran Wang, and Luc~Van Gool.
\newblock {MIC:} masked image consistency for context-enhanced domain adaptation.
\newblock In {\em Proceedings of the {IEEE/CVF} Conference on Computer Vision and Pattern Recognition (CVPR)}, pages 11721--11732, 2023.

\bibitem{DBLP:conf/cvpr/HuGZDW18}
Han Hu, Jiayuan Gu, Zheng Zhang, Jifeng Dai, and Yichen Wei.
\newblock Relation networks for object detection.
\newblock In {\em Proceedings of the IEEE/CVF Conference on Computer Vision and Pattern Recognition (CVPR)}, pages 3588--3597, 2018.

\bibitem{DBLP:conf/cvpr/HuangL0HCL23}
Duojun Huang, Jichang Li, Weikai Chen, Junshi Huang, Zhenhua Chai, and Guanbin Li.
\newblock Divide and adapt: Active domain adaptation via customized learning.
\newblock In {\em Proceedings of the {IEEE/CVF} Conference on Computer Vision and Pattern Recognition (CVPR)}, pages 7651--7660, 2023.

\bibitem{DBLP:conf/nips/HuangGXL21}
Jiaxing Huang, Dayan Guan, Aoran Xiao, and Shijian Lu.
\newblock Model adaptation: Historical contrastive learning for unsupervised domain adaptation without source data.
\newblock In {\em Proceedings of the Annual Conference on Neural Information Processing Systems (NeurIPS)}, pages 3635--3649, 2021.

\bibitem{DBLP:conf/iclr/Hwang0SY24}
Uiwon Hwang, Jonghyun Lee, Juhyeon Shin, and Sungroh Yoon.
\newblock {SF(DA)2:} source-free domain adaptation through the lens of data augmentation.
\newblock In {\em Proceedings of the International Conference on Learning Representations (ICLR)}, 2024.

\bibitem{DBLP:conf/icra/JeonSM24}
Minsik Jeon, Junwon Seo, and Jihong Min.
\newblock {DA-RAW:} domain adaptive object detection for real-world adverse weather conditions.
\newblock In {\em Proceedings of the {IEEE} International Conference on Robotics and Automation (ICRA)}, pages 2013--2020, 2024.

\bibitem{DBLP:journals/corr/Johnson-Roberson16}
Matthew Johnson{-}Roberson, Charles Barto, Rounak Mehta, Sharath~Nittur Sridhar, and Ram Vasudevan.
\newblock Driving in the matrix: Can virtual worlds replace human-generated annotations for real world tasks?
\newblock {\em CoRR}, abs/1610.01983, 2016.

\bibitem{DBLP:journals/tmlr/KayHSDYPBH25}
Justin Kay, Timm Haucke, Suzanne Stathatos, Siqi Deng, Erik Young, Pietro Perona, Sara Beery, and Grant~Van Horn.
\newblock Align and distill: Unifying and improving domain adaptive object detection.
\newblock {\em Trans. Mach. Learn. Res.}, 2025, 2025.

\bibitem{DBLP:conf/eccv/KhanhNPTJ24}
Trinh Le~Ba Khanh, Huy{-}Hung Nguyen, Long~Hoang Pham, Duong~Nguyen{-}Ngoc Tran, and Jae~Wook Jeon.
\newblock Dynamic retraining-updating mean teacher for source-free object detection.
\newblock In {\em Proceedings of the European Conference on Computer Vision (ECCV)}, pages 328--344, 2024.

\bibitem{DBLP:conf/iccv/KodirovXFG15}
Elyor Kodirov, Tao Xiang, Zhen{-}Yong Fu, and Shaogang Gong.
\newblock Unsupervised domain adaptation for zero-shot learning.
\newblock In {\em Proceedings of the IEEE/CVF Conference on Computer Vision (ICCV)}, pages 2452--2460, 2015.

\bibitem{DBLP:journals/corr/abs-2506-15742}
Black~Forest Labs, Stephen Batifol, Andreas Blattmann, Frederic Boesel, Saksham Consul, Cyril Diagne, Tim Dockhorn, Jack English, Zion English, Patrick Esser, Sumith Kulal, Kyle Lacey, Yam Levi, Cheng Li, Dominik Lorenz, Jonas M{\"{u}}ller, Dustin Podell, Robin Rombach, Harry Saini, Axel Sauer, and Luke Smith.
\newblock {FLUX.1} kontext: Flow matching for in-context image generation and editing in latent space.
\newblock {\em CoRR}, abs/2506.15742, 2025.

\bibitem{DBLP:conf/iclr/LaineA17}
Samuli Laine and Timo Aila.
\newblock Temporal ensembling for semi-supervised learning.
\newblock In {\em Proceedings of the International Conference on Learning Representations (ICLR)}, 2017.

\bibitem{DBLP:conf/cvpr/LeeFP13}
Gim~Hee Lee, Friedrich Fraundorfer, and Marc Pollefeys.
\newblock Motion estimation for self-driving cars with a generalized camera.
\newblock In {\em Proceedings of the IEEE/CVF Conference on Computer Vision and Pattern Recognition (CVPR)}, pages 2746--2753, 2013.

\bibitem{DBLP:conf/cvpr/Li0ZZX22}
Shuaifeng Li, Mao Ye, Xiatian Zhu, Lihua Zhou, and Lin Xiong.
\newblock Source-free object detection by learning to overlook domain style.
\newblock In {\em Proceedings of the {IEEE/CVF} Conference on Computer Vision and Pattern Recognition (CVPR)}, pages 8004--8013, 2022.

\bibitem{DBLP:conf/cvpr/LiDMLCWHKV22}
Yu{-}Jhe Li, Xiaoliang Dai, Chih{-}Yao Ma, Yen{-}Cheng Liu, Kan Chen, Bichen Wu, Zijian He, Kris Kitani, and Peter Vajda.
\newblock Cross-domain adaptive teacher for object detection.
\newblock In {\em Proceedings of the {IEEE/CVF} Conference on Computer Vision and Pattern Recognition (CVPR)}, pages 7571--7580, 2022.

\bibitem{DBLP:conf/icml/LiangHF20}
Jian Liang, Dapeng Hu, and Jiashi Feng.
\newblock Do we really need to access the source data? source hypothesis transfer for unsupervised domain adaptation.
\newblock In {\em Proceedings of the International Conference on Machine Learning (ICML)}, pages 6028--6039, 2020.

\bibitem{DBLP:conf/iccv/0004LSY23}
Qipeng Liu, Luojun Lin, Zhifeng Shen, and Zhifeng Yang.
\newblock Periodically exchange teacher-student for source-free object detection.
\newblock In {\em Proceedings of the {IEEE/CVF} International Conference on Computer Vision (ICCV)}, pages 6391--6401, 2023.

\bibitem{DBLP:conf/eccv/LuKBL16}
Cewu Lu, Ranjay Krishna, Michael~S. Bernstein, and Li~Fei{-}Fei.
\newblock Visual relationship detection with language priors.
\newblock In {\em Proceedings of the European Conference on Computer Vision (ECCV)}, pages 852--869, 2016.

\bibitem{DBLP:conf/cvpr/MaquedaL0GS18}
Ana~I. Maqueda, Antonio Loquercio, Guillermo Gallego, Narciso Garc{\'{\i}}a, and Davide Scaramuzza.
\newblock Event-based vision meets deep learning on steering prediction for self-driving cars.
\newblock In {\em Proceedings of the IEEE/CVF Conference on Computer Vision and Pattern Recognition (CVPR)}, pages 5419--5427, 2018.

\bibitem{DBLP:journals/pami/OzaSVP24}
Poojan Oza, Vishwanath~A. Sindagi, Vibashan VS, and Vishal~M. Patel.
\newblock Unsupervised domain adaptation of object detectors: {A} survey.
\newblock {\em {IEEE} Trans. Pattern Anal. Mach. Intell.}, 46(6):4018--4040, 2024.

\bibitem{DBLP:conf/nips/RenHGS15}
Shaoqing Ren, Kaiming He, Ross~B. Girshick, and Jian Sun.
\newblock Faster {R-CNN:} towards real-time object detection with region proposal networks.
\newblock In {\em Proceedings of the Annual Conference on Neural Information Processing Systems (NeurIPS)}, pages 91--99, 2015.

\bibitem{DBLP:conf/nips/TarvainenV17}
Antti Tarvainen and Harri Valpola.
\newblock Mean teachers are better role models: Weight-averaged consistency targets improve semi-supervised deep learning results.
\newblock In {\em Proceedings of the Annual Conference on Neural Information Processing Systems (NeurIPS)}, pages 1195--1204, 2017.

\bibitem{DBLP:conf/cvpr/VSO023}
Vibashan VS, Poojan Oza, and Vishal~M. Patel.
\newblock Instance relation graph guided source-free domain adaptive object detection.
\newblock In {\em Proceedings of the {IEEE/CVF} Conference on Computer Vision and Pattern Recognition (CVPR)}, pages 3520--3530, 2023.

\bibitem{DBLP:conf/eccv/WangG18}
Xiaolong Wang and Abhinav Gupta.
\newblock Videos as space-time region graphs.
\newblock In {\em Proceedings of the European Conference on Computer Vision (ECCV)}, pages 413--431, 2018.

\bibitem{wu2019detectron2}
Yuxin Wu, Alexander Kirillov, Francisco Massa, Wan-Yen Lo, and Ross Girshick.
\newblock Detectron2.
\newblock \url{https://github.com/facebookresearch/detectron2}, 2019.

\bibitem{DBLP:journals/ijis/XiongYZGLZ21}
Lin Xiong, Mao Ye, Dan Zhang, Yan Gan, Xue Li, and Yingying Zhu.
\newblock Source data-free domain adaptation of object detector through domain-specific perturbation.
\newblock {\em Int. J. Intell. Syst.}, 36(8):3746--3766, 2021.

\bibitem{DBLP:conf/cvpr/XuZCF17}
Danfei Xu, Yuke Zhu, Christopher~B. Choy, and Li~Fei{-}Fei.
\newblock Scene graph generation by iterative message passing.
\newblock In {\em Proceedings of the IEEE/CVF Conference on Computer Vision and Pattern Recognition (CVPR)}, pages 3097--3106, 2017.

\bibitem{DBLP:conf/iccv/YangW0HJ21}
Shiqi Yang, Yaxing Wang, Joost van~de Weijer, Luis Herranz, and Shangling Jui.
\newblock Generalized source-free domain adaptation.
\newblock In {\em Proceedings of the IEEE/CVF Conference on Computer Vision (ICCV)}, pages 8958--8967, 2021.

\bibitem{DBLP:conf/eccv/YoonKKPJS24}
Ilhoon Yoon, Hyeongjun Kwon, Jin Kim, Junyoung Park, Hyunsung Jang, and Kwanghoon Sohn.
\newblock Enhancing source-free domain adaptive object detection with low-confidence pseudo label distillation.
\newblock In {\em Proceedings of the European Conference on Computer Vision (ECCV)}, volume 15142, pages 337--353, 2024.

\bibitem{DBLP:conf/mm/YouLZCH21}
Fuming You, Jingjing Li, Lei Zhu, Zhi Chen, and Zi~Huang.
\newblock Domain adaptive semantic segmentation without source data.
\newblock In {\em Proceedings of the {ACM} Multimedia Conference (MM)}, pages 3293--3302, 2021.

\bibitem{DBLP:journals/corr/abs-2202-02802}
Yixin Zhang, Junjie Li, and Zilei Wang.
\newblock Low-confidence samples matter for domain adaptation.
\newblock {\em CoRR}, abs/2202.02802, 2022.

\bibitem{DBLP:conf/nips/ZhangCCLLLL22}
Ziyi Zhang, Weikai Chen, Hui Cheng, Zhen Li, Siyuan Li, Liang Lin, and Guanbin Li.
\newblock Divide and contrast: Source-free domain adaptation via adaptive contrastive learning.
\newblock In {\em Proceedings of the Annual Conference on Neural Information Processing Systems (NeurIPS)}, 2022.

\bibitem{DBLP:conf/iccv/ZhaoWCLYPL23}
Zijing Zhao, Sitong Wei, Qingchao Chen, Dehui Li, Yifan Yang, Yuxin Peng, and Yang Liu.
\newblock Masked retraining teacher-student framework for domain adaptive object detection.
\newblock In {\em Proceedings of the {IEEE/CVF} International Conference on Computer Vision (ICCV)}, pages 18993--19003, 2023.

\end{thebibliography}
